\documentclass[10pt,twocolumn,letterpaper]{article}

\usepackage{cvpr}              
\usepackage{graphicx}
\usepackage{multirow}
\usepackage{bbm}
\usepackage{colortbl}
\usepackage{ulem}
\usepackage{algorithm}
\usepackage{algorithmicx}
\usepackage{pifont}
\definecolor{cvprblue}{rgb}{0.21,0.49,0.74}
\usepackage[pagebackref,breaklinks,colorlinks,allcolors=cvprblue]{hyperref}

\def\paperID{3891} 
\def\confName{CVPR}
\def\confYear{2025}

\title{SeCap: Self-Calibrating and Adaptive Prompts for\\ Cross-view Person Re-Identification in Aerial-Ground Networks}

\author{Shining Wang$^{1,2,4}$\footnotemark[1], Yunlong Wang$^{1,2,4}$\footnotemark[1], Ruiqi Wu$^{1,2,4}$, 
Bingliang Jiao$^{1,2,4}$, Wenxuan Wang$^{1,3,4}$\footnotemark[2], Peng Wang$^{1,2,4}$
\\
$^1$School of Computer Science, Northwestern Polytechnical University, China.\\
$^2$Ningbo Institute, Northwestern Polytechnical University, China.\\
$^3$Shenzhen Research Institute, Northwestern Polytechnical University, China.\\
$^4$National Engineering Laboratory for integrated Aero-Space-Ground-Ocean, Xi'an, Shaanxi, China.\\
{\tt\small \{wangshining, wangyunlong2019, wurq, bingliang.jiao\}@mail.nwpu.edu.cn
\tt\small \{wxwang, peng.wang\}@nwpu.edu.cn
}}

\begin{document}
\maketitle
\renewcommand\thefootnote{\fnsymbol{footnote}}
\footnotetext[1]{These authors contributed equally to this work.} 
\footnotetext[2]{Corresponding authors.} 
\begin{abstract}
When discussing the Aerial-Ground Person Re-identification (AGPReID) task, we face the main challenge of the significant appearance variations caused by different viewpoints, making identity matching difficult. To address this issue, previous methods attempt to reduce the differences between viewpoints by critical attributes and decoupling the viewpoints. While these methods can mitigate viewpoint differences to some extent, they still face two main issues: (1) difficulty in handling viewpoint diversity and (2) neglect of the contribution of local features. To effectively address these challenges, we design and implement the Self-Calibrating and Adaptive Prompt (SeCap) method for the AGPReID task. The core of this framework relies on the Prompt Re-calibration Module (PRM), which adaptively re-calibrates prompts based on the input. Combined with the Local Feature Refinement Module (LFRM), SeCap can extract view-invariant features from local features for AGPReID. Meanwhile, given the current scarcity of datasets in the AGPReID field, we further contribute two real-world Large-scale Aerial-Ground Person Re-Identification datasets, LAGPeR and G2APS-ReID. The former is collected and annotated by us independently, covering $4,231$ unique identities and containing $63,841$ high-quality images; the latter is reconstructed from the person search dataset G2APS. Through extensive experiments on AGPReID datasets, we demonstrate that SeCap is a feasible and effective solution for the AGPReID task. The datasets and source code available on \href{https://github.com/wangshining681/SeCap-AGPReID}{https://github.com/wangshining681/SeCap-AGPReID}.
\end{abstract}    

\begin{figure}[t] 
    \centering 
    \includegraphics[width=\linewidth]{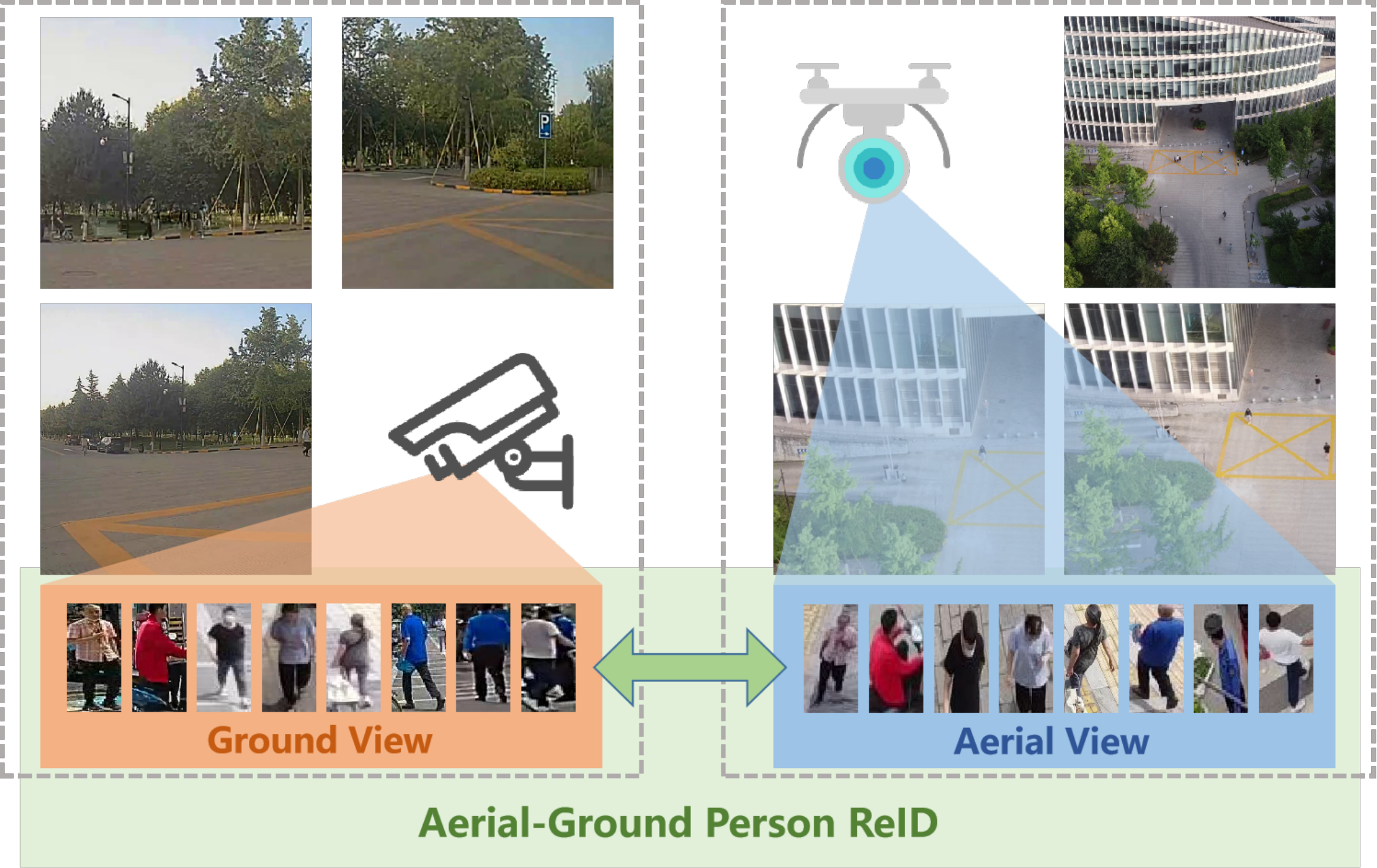}
    \caption{Aerial View and Ground View exhibit significant appearance variation due to notable differences in views. This variation poses substantial challenges for cross-view image matching.}
    \label{fig:intro}
\end{figure}

\begin{table*}[ht]
\begin{center}
   \caption{Statistical comparisons with existing datasets, including view-homogeneous (ground or aerial) and view-heterogeneous (ground and aerial) ReID.} 
    \label{tab:dataset}
\end{center}
\vspace{-0.5cm}
\renewcommand\arraystretch{1.2}
\centering
\setlength{\tabcolsep}{1.5mm}{
    \begin{tabular}{lcccccc} \hline 
         \multicolumn{1}{c}{\textsc{Dataset}} &  \textsc{View} &  \textsc{Source} &  \textsc{\#Identity} &  \textsc{\#Camera} & \textsc{\#Image} & \textsc{Height}\\ \hline \hline
         Market1501~\cite{zheng2015person}&Ground&Real&1,501 &16& 32,668& \textless 10m \\
         DukeMTMC-reID~\cite{zheng2017unlabeled}&Ground&Real& 1,404& 8& 36,411 & \textless 10m\\ 
         PRAI1581~\cite{zhang2020person}&  Aerial&  Real&  1,581&  2 &  39,461 &20$\sim$60m\\ 
         UAVHuman~\cite{li2021uav}&  Aerial&  Real&  1,144&  1 &   41,290 & 2-8m \\ \hline
         AG-ReID.v1~\cite{nguyen2023aerial} & Aerial-Ground & Real & 388 & 2(1A+1G) & 21,893 &15$\sim$45m\\ 
        AG-ReID.v2~\cite{nguyen2024ag} &	Aerial-Ground-Wearable & Real &	1,615 &	3(1A+1G+1W) & 100,502 & 15$\sim$45m \\ 
        CARGO~\cite{zhang2024view}&	Aerial-Ground&	Synthetic&	5,000 &	13(5A+8G) &	108,563 & 5$\sim$75m \\ 
        \hline
        \textbf{G2APS-ReID(Ours)}&	\textbf{Aerial-Ground}&	\textbf{Real}&	\textbf{2,788}&	\textbf{2(1A+1G)}&	\textbf{200,864} & \textbf{20$\sim$60m} \\
        \textbf{LAGPeR(Ours)}&	\textbf{Aerial-Ground}&	\textbf{Real}&	\textbf{4,231}&	\textbf{21(7A+14G)}&	\textbf{63,841} & \textbf{20$\sim$60m} \\ \hline
    \end{tabular}
}
\end{table*}

\section{Introduction}
\label{sec:intro}

Person re-identification (ReID), as the cornerstone of intelligent surveillance systems, fundamentally relies on accurately identifying individuals across different camera viewpoints and complex environmental changes~\cite{lee2023camera, li2021weperson, qian2020long,  zhang2024separable,  qian2018pose,wang2024exploring}. However, traditional ReID methods are often limited to the same view, such as ground view~\cite{ye2021deep, zhang_alignedreid_2018,sun2018beyond, zhang_magic_2024, zhang2024uncertainty,qian2019leader} or aerial view~\cite{zhang2020person, li2021uav, chen2022rotation,khaldi2024unsupervised,wang2024rotation}, and fail to adequately address the challenges posed by extreme visual transformations and the integration of complementary information in cross-view scenarios (e.g., combining ground and aerial cameras). These challenges are more prevalent in real-world applications, thereby introducing the problem of the Aerial-Ground Person Re-Identification (AGPReID).

As shown in Fig.~\ref{fig:intro}, unlike traditional ReID tasks, the query and gallery sets in AGPReID are captured separately by aerial-view or ground-view cameras~\cite{nguyen2023aerial,nguyen2024ag,zhang2024view}, respectively. This results in significant variations in the compared person images, making identity matching more challenging~\cite{yan_bv-person_2021,liu_scale-invariant_2024}. To address this issue, AG-ReID~\cite{nguyen2023aerial,nguyen2024ag} utilizes identity attributes to extract view-invariant information, partially solving the cross-view matching problem. Additionally, VDT~\cite{zhang2024view} designs the view decoupling transformer based on ViT~\cite{liu2021viT}, using a hierarchical decoupling mechanism to extract view-invariant features effectively.


Although existing methods can partially solve the cross-view matching problem, they still suffer from over-reliance on attributes, insufficient adaptation to multiple views, etc. Specifically, there are two main issues with current methods: (1) \textit{Difficulty in Handling Viewpoint Diversity}: In AGPReID tasks, the variability of viewpoints renders decoupling methods within a single viewpoint insufficient for all viewpoints.  (2) \textit{Neglect of Local Features}: Due to the steep downward viewing angles of drones, various parts of the body are often not fully exposed in the images, especially when person are occluded~\cite{he2021transreid,miao_pose-guided_2019,sun2018beyond,ning2024enhancement}. AG-ReID~\cite{nguyen2023aerial} extracts cross-view information by attributes, which essentially rely on view-invariant local features. However, the attribute labels hinder generalization. Therefore, we must consider learning view-invariant local features from different viewpoints without dependency on labels.


To overcome these challenges, this paper proposes an AGPReID framework named SeCap, which self-calibrates and adaptively generates prompts based on the inputs for cross-view person re-identification. This framework adopts an encoder-decoder transformer architecture. The encoder employs the View Decoupling Transformer (VDT) for viewpoint decoupling, while the decoder further decodes local features using the view-invariant features. Specifically, the decoder comprises the Prompt Re-calibration Module (PRM) and the Local Feature Refinement Module (LPRM). To address the challenge of viewpoint diversity (limitation 1), we design the PRM to re-calibrate prompts based on the input adaptively. It dynamically generates and self-calibrates prompts that closely align with the current viewpoint, thus adapt to different viewpoints. To fully leverage the role of local features (limitation 2), we design the LPRM for local feature refinement. This module uses re-calibrated prompts and employs the to-way attention mechanism to synchronously update various features, thereby learning view-invariant information from local features.


Furthermore, as shown in Tab.~\ref{tab:dataset}, due to the scarcity of datasets for the AGPReID task, we contribute two real-world large-scale aerial-ground person re-identification datasets, LAGPeR and G2APS-ReID. The LAGPeR dataset, which is independently collected, annotated, and partitioned by us, is gathered from campus scenes using a combination of seven drone cameras and fourteen ground cameras, capturing a total of $4,231$ unique identities across $63,841$ images. This dataset fully considers the impacts of occlusion, lighting, and cross-domain variations that may be encountered in real-world applications. To further expand the AGPReID dataset, we also reconstruct the G2APS-ReID from the person search dataset G2APS~\cite{zhang2023ground}. For the comprehensive evaluation of various AGPReID methods, we meticulously design the evaluation settings that account for the retrieval demands across both ground and aerial viewpoints, thereby thoroughly evaluating the models' effectiveness in handling cross-view re-identification tasks. In summary, the main contributions of this paper include:
\begin{itemize}
\item Introducing an innovative AGPReID method called SeCap, adaptively re-calibrate prompts to match the current view based on the input, significantly enhancing the AGPReID performance.
\item Contribute two real-world large-scale AGPReID datasets, LAGPeR and G2APS-ReID, which provide significant data support for research in the AGPReID task.
\item Conduct extensive experiments on AGPReID datasets, demonstrating our SeCap's superiority in AGPReID tasks and achieving state-of-the-art (SOTA) performance.
\end{itemize}

\section{Related Work}
\label{sec:related_work}
\subsection{View-Invariant Person Re-Identification}
Research on view-invariant ReID typically focuses on two main categories: ground-view~\cite{chen2024region, zheng2015person, zheng2017unlabeled,huang2024meta} and aerial-view perspectives~\cite{zhang2020person, li2021uav, chen2022rotation}. 
Existing studies contribute a plethora of ground-view datasets, such as Market1501~\cite{zheng2015person} and MSMT17~\cite{wei2018person}, which significantly advance the field. In recent years, with the maturation and widespread application of Transformer architectures, Vision Transformer (ViT) models~\cite{liu2021viT} gradually emerged in the ReID domain~\cite{he2021transreid}, become the mainstream backbone for feature extraction. Against this backdrop, as an emerging technological paradigm, prompt learning is widely applied in various ReID tasks~\cite{he2024instruct,li2023clip}, demonstrating its ability to retain the inherent knowledge of the backbone model while adapting it to various tasks~\cite{wu2024enhancing,he2024instruct,yang2024robust,qin2024noisy}. This inspires us to use prompt learning to solve the AGPReID problem. Overall, compared to view-invariant ReID, research on cross-view ReID is relatively sparse, especially concerning the significant appearance changes caused by aerial and ground-view variations. Therefore, our work focuses on the aerial-ground cross-view person ReID task, aiming to find an effective method to address this challenge.

\subsection{Aerial-Ground Cross-View ReID}

The core challenge of the aerial-ground person re-identification (AGPReID) lies in the significant appearance variations caused by different viewpoints, making identity matching difficult. To address this challenge, AG-ReID~\cite{nguyen2023aerial} solves the cross-view matching problem by leveraging identity attributes, making individuals with similar attributes more likely to be identified as the same person. Meanwhile, AG-ReIDv2~\cite{nguyen2024ag} introduces the Elevated-View Attention Stream to fully utilize the invariant local information between ground and aerial view for identity discrimination, which also demonstrates the importance of local features in AGPReID tasks. However, these two works rely on attributes and do not consider the decoupling of view-invariant information within local features, limiting the scalability of the method. Another approach aligns feature spaces from different viewpoints, as in VDT~\cite{zhang2024view}, which designs the view-decoupling transformer based on ViT~\cite{liu2021viT}, effectively extracting view-invariant features through a hierarchical decoupling mechanism. Although this method can partially decouple view-related information, the diversity of viewpoints in AGPReID may lead to over-separation of potentially useful information within view-invariant features, and even result in the loss of inherent knowledge of the backbone network. Different from existing works, our proposed framework, SeCap, can adaptively adjust prompts based on inputs to generate prompts suitable for different viewpoints, effectively separating view-invariant features. 

\subsection{Aerial-Ground ReID Datasets}
In the AGPReID field, the scarcity and limitations of datasets become the key factors constraining research progress. Compared to mature view-invariant datasets~\cite{zheng2015person,zheng2017unlabeled}, AGPReID datasets are scarce in number and far from meeting research needs in terms of real-world data scale. For instance, the number of identity IDs in AG-ReID.v1~\cite{nguyen2023aerial} datasets is far lower than in-ground ReID datasets like Market1501~\cite{zheng2015person} and DukeMTMC-reID~\cite{zheng2017unlabeled}, highlighting the lack of data resources in this field. Although AG-ReID.v2~\cite{nguyen2024ag} attempts to address this shortfall by expanding the dataset, the large number of wearable device images included in the new data does not substantially enhance coverage of the differences between aerial and ground views, affecting its practical value. Additionally, synthetic datasets like CARGO~\cite{zhang2024view}, while approaching real datasets in scale, lack real-world complexity, limiting model performance in actual scenarios. Therefore, developing the larger-scale aerial-ground cross-view ReID dataset, and designing efficient algorithms to utilize these datasets effectively, become urgent problems to be solved in this field. To overcome this problem, we contribute two real-world large-scale AGPReID datasets: LAGPeR and G2APS-ReID. 

\begin{figure*}[ht] 
    \centering 
    \includegraphics[width=\textwidth]{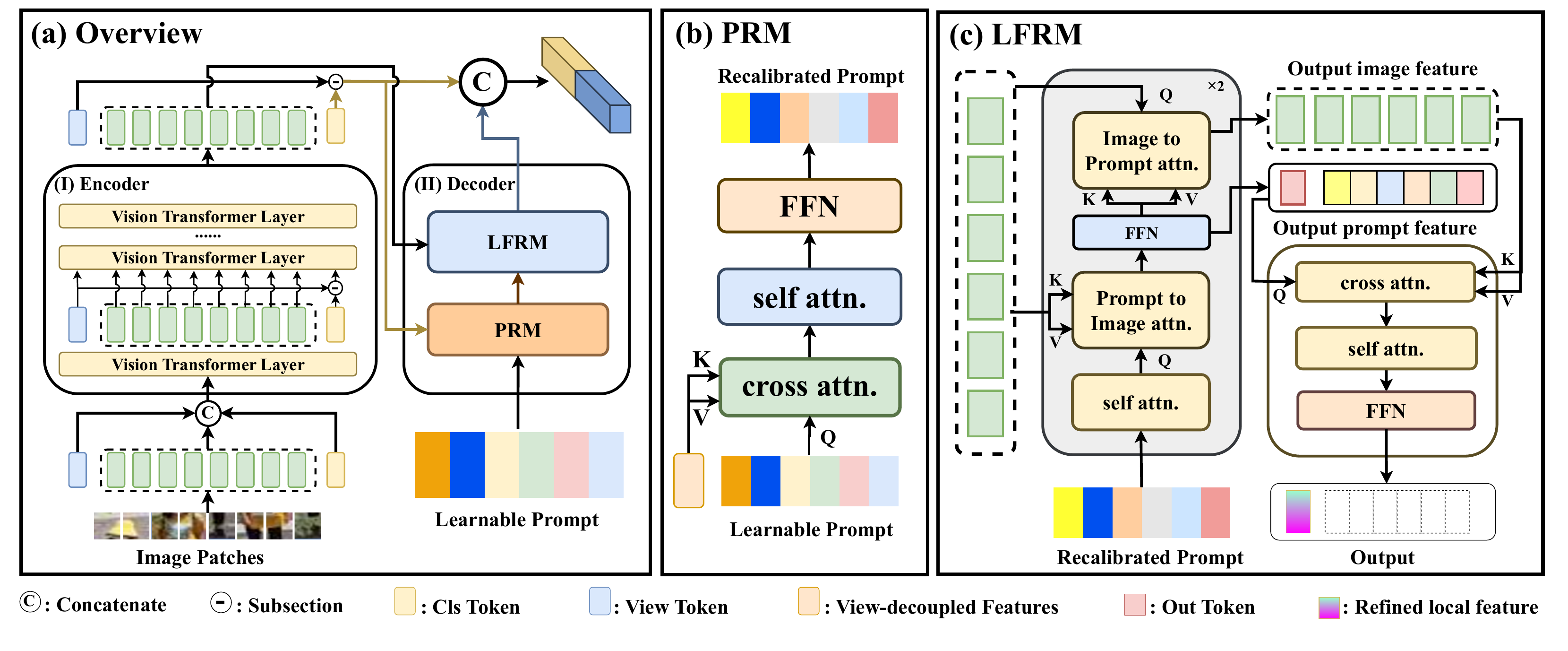}
    \caption{
    (a) The architecture of the proposed SeCap. The key component is an encoder-decoder transformer. The encoder extracts the visual features of the picture and decouples the viewpoints. The decoder re-calibrates prompts through the current viewpoint information and decodes the local features using the re-calibrated prompts. (b) The Prompt Re-calibration Module (PRM) adaptively generates and re-calibrates prompts for different viewpoints according to view-invariant features. (c) The Local Feature Refinement Module (LFRM) finely decodes discriminative features from the local features using the re-calibrated prompts in PRM.
    }
    \label{fig:framework}
\end{figure*}

\section{Method}
\subsection{Overview} 
The overall framework of SeCap, as illustrated in Fig.~\ref{fig:framework} (a), adopts an encoder-decoder transformer architecture. The encoder is the view decoupling transformer (VDT)~\cite{zhang2024view}. In contrast to the conventional ViT~\cite{liu2021viT}, our approach incorporates the \textbf{View} token and performs hierarchical decoupling of the \textbf{Cls} token at each layer, effectively segregating view-related and view-invariant features within the \textbf{Cls} token, while extracting local features from the input. 
The decoder comprises the Prompt Re-calibration Module (\textbf{PRM}) and the Local Feature Refinement Module (\textbf{LFRM}). The \textbf{PRM} adaptively generates and re-calibrates prompts for different viewpoints based on the current viewpoint information. Concurrently, the \textbf{LFRM} utilizes the re-calibrated prompts from the \textbf{PRM} to decode the local features. The overall framework can be described as follows:
\begin{equation}
    \begin{aligned}
[\text{\textbf{Cls}}, \text{\textbf{View}}, X_{\text{local}}] &= \text{VDT}([\text{\textbf{CLS}}, \text{\textbf{View}}, [\text{tokenization}(\textbf{X})]]) \\
X_{\text{inv}} &= \text{\textbf{Cls}} - \text{\textbf{View}} \\
X_{\text{local}} &= \text{\textbf{LFRM}}(X_{\text{local}}, \text{\textbf{PRM}}(\text{\textbf{Prompt}}, X_{\text{inv}})) \\
\text{\textbf{Out}} &= [X_{\text{inv}}, X_{\text{local}}]
    \end{aligned}
\end{equation}
where VDT is the View Decoupling Transformer, $\text{\textbf{Cls}}$ and $\text{\textbf{View}}$ are the Class and View token, $\text{tokenization}(\textbf{X})$ refers to the process of converting the inputs $\textbf{X}$ into tokens, $X_{\text{inv}}$ represents the view-invariant features, $\text{\textbf{Prompt}}$ denotes learnable prompts, $X_{\text{local}}$ signifies the local features of the input data, and $\text{\textbf{Out}}$ is the final output.

\subsection{Prompt Re-calibration Module}
The Prompt Re-calibration Module (\textbf{PRM}) is designed based on the Transformer Decoder architecture~\cite{carion2020end}, aiming at adaptively re-calibrating prompts suitable for different viewpoints. Specifically, this module initializes and maintains a set of prompts with learnable vectors $\text{\textbf{Prompt}} = \mathrm{[{Prompt}_1, {Prompt}_2,\dots,{Prompt}_\textbf{L}]}$, where $\text{\textbf{L}}$ is the hyperparameter that denotes the prompt length. 

As illustrated in Fig.~\ref{fig:framework} (b), the module initially incorporates the view-invariant features into prompts via cross-attention, enhancing prompts' focus on view-invariant information within visual features. 
Subsequently, the self-attention mechanism is employed to amalgamate and re-calibrate the information of each prompt within the prompts sequence, ensuring comprehensive integration of the view-invariant information.
Finally, the Feed-Forward Network ($\text{FFN}$) is applied to produce the re-calibrated prompts tailored to the view-invariant information. The Prompt Re-calibration Module can be described as follows:
\begin{equation}
    \begin{aligned}
    \text{P}_{\text{re}} = \text{FFN}(\text{SA}(\text{CA}(\text{\textbf{Prompt}}, X_{\text{inv}}, X_{\text{inv}}))) + \text{\textbf{Prompt}}
    \end{aligned}
\end{equation}
where $X_{\text{inv}}$ represents the view-invariant features extracted from the backbone, $\text{\textbf{Prompt}}$ denotes learnable prompts, $\text{CA}$ is the cross-attention mechanism that aligns prompts with the view-invariant features, $\text{SA}$ is the self-attention mechanism that enables the model to weigh the importance of different parts of the input, $\text{FFN}$ stands for the Feed-Forward Network, and $\text{P}_{\text{re}}$ is the re-calibrated prompts.

\subsection{Local Feature Refinement Module}
The Local Feature Refinement Module (\textbf{LFRM}) is the Transformer-based decoder, as illustrated in Fig~\ref{fig:framework} (c). It extracts the view-invariant features of the local features $\text{F}_{\text{local}}$, using the re-calibrated prompts $\text{P}_{\text{re}}$ from the \textbf{PRM}. The re-calibrated prompts integrate view-invariant information from the global features, enabling the \textbf{LFRM} to decode view-invariant features of the local features, thereby aligning the local features with the global features. \\
Specifically, the \textbf{LFRM} consists of the two-way attention module and the feature fusion module. The two-way attention module employs both self-attention and cross-attention mechanisms in both prompt-to-image encoding and image-to-prompt encoding directions, to dynamically update and enhance all feature representations. Through the two-way attention module, the \textbf{LFRM} efficiently integrates and updates the visual information of the local features and view-invariant information of the re-calibrated prompts. The two-way attention module can be described as follows:
\begin{equation}
    \begin{aligned}
    \text{F}_\text{P} &= \text{FFN}(\text{CA}(\text{SA}(\text{P}_{\text{re}}), \text{F}_{\text{local}}, \text{F}_{\text{local}})) + \text{P}_{\text{re}}  \\
    \text{F}_\text{I} &= \text{CA}(\text{F}_{\text{local}}, \text{F}_\text{P}, \text{F}_\text{P}) + \text{F}_{\text{local}}
    \end{aligned}
\end{equation}
where $\text{F}_{\text{P}}$ represents the prompt features output by the two-way attention module, $\text{F}_{\text{I}}$ denotes the image features output by the two-way attention module, $\text{FFN}$ stands for the Feed-Forward Network, $\text{CA}$ is the cross-attention mechanism, $\text{SA}$ is the self-attention mechanism, $\text{F}_{\text{local}}$ signifies the local features, and $\text{P}_{\text{re}}$ refers to the re-calibrated prompts. \\
To maintain the light weight of the decoder, we only stack two two-way attention blocks in \textbf{LFRM}. Additionally, the feature fusion module employs cross-attention and self-attention to further integrate the image features and prompt features output by the two-way attention module, thereby decoding view-invariant features of the local features. The feature fusion module can be described as follows:
\begin{equation}
    \begin{aligned}
    [\text{\textbf{Out}},  \_] = \text{FFN}(\text{SA}(\text{CA}([\text{\textbf{Out}}, \text{F}_\text{P}], \text{F}_\text{I}, \text{F}_\text{I})))
    \end{aligned}
\end{equation}
where $\text{\textbf{Out}}$ is output token, which integrates the final output, $\text{F}_\text{P}$ is the prompt features, $\text{F}_\text{I}$ is the image features.

\subsection{Optimization}
\label{sec:optimization}
In addition to the commonly employ the ID classification loss and the Triplet loss in ReID tasks, the loss functions in our model are further enhanced with the view classification loss and the orthogonality loss.\\
\textbf{View Classification Loss}: To achieve the decoupling of view-related features, we utilize a view classifier to guide the extraction of view-related features and employ the view classification loss for constraint. The loss function for this process can be formalized as follows:
\begin{equation}
    \begin{aligned}
\mathcal{L}_{\text{view}} = -\sum_{i=1}^{N} y_i \log(p_i)
    \end{aligned}
\end{equation}
where $y_i$ is the ground truth label of view for the $i$-th sample, indicating the actual view class, $p_i$ is the predicted probability of the $i$-th sample belonging to the correct view class, and $N$ is the total number of samples in the dataset.\\
\textbf{Orthogonality Loss:} To quantify the effectiveness of view decoupling, we introduce the orthogonality loss to ensure thorough decoupling. The specific expression of the orthogonality loss is as follows:
\begin{equation}
    \begin{aligned}
\mathcal{L}_{\text{orth}} = \sum_{i=1}^{d} | \langle \mathbf{inv}_i, \mathbf{v}_i \rangle |
    \end{aligned}
\end{equation}
where $|\langle \cdot, \cdot \rangle|$ denotes the absolute value of the dot product between two features, which is used to measure their linear correlation. $d$ represents the dimensionality of the features. $\mathbf{inv}_i$ and $\mathbf{v}_i$ refer to the $i$-th dimension of the view-invariant and view-related features.\\
\textbf{Overall Loss:} We apply the ID classification loss and the Triplet loss to the global and local features to guide the model's learning. Due to the significant difference in the number of categories between the view classifier and the ID classifier (in the thousands), the view classification loss is scaled by a small coefficient $\lambda$ to balance the difficulty between view and ID classification. The impact of this coefficient is analyzed in \textbf{Supplementary Material}. The overall optimization objective can be summarized as follows:
\begin{equation}
    \begin{aligned}
        \mathcal{L} &= \alpha (\mathcal{L}_{ID}^{global} + \mathcal{L}_{Tri}^{global}) \\
                    &\quad + \beta (\mathcal{L}_{ID}^{local} + \mathcal{L}_{Tri}^{local}) \\
                    &\quad + \lambda (\mathcal{L}_v + \mathcal{L}_{\rho}),
    \end{aligned}
\end{equation}
where $\lambda$ is the hyperparameter used to balance the different difficulties of view classification and ID classification, $\alpha$ and $\beta$ are also the hyperparameter to balance the optimization objectives of global and local features.

\begin{table*}[tp]
\begin{center}
   \caption{Experimental setup and data division of the LAGPeR and G2APS-ReID datasets.} 
   \label{tab:split}
\end{center}
\vspace{-0.5cm}
\renewcommand\arraystretch{1.2}
\centering
\setlength{\tabcolsep}{3.5mm}{
    \begin{tabular}{ccc|ccc|ccc} \hline
    \multirow{2}{*}{\textsc{Setting}} &
    \multirow{2}{*}{\textsc{Subset}} &
    \multirow{2}{*}{\#View.}&
    \multicolumn{3}{c|}{LAGPeR}&
    \multicolumn{3}{c}{G2APS-ReID}\\	
    &&&\#Cam    &\#IDs    &\#Images &\#Cam    &\#IDs    &\#Images \\
    \hline \hline
    - & Train & Aerial+Ground & 12 & 2,708 & 40,770 & 2 & 1,569 & 100,871 \\
    \hline
    \multirow{2}*{$A \rightarrow G$} & Query & Aerial & 3 & 1,523 & 3,046 & 1 & 1,219 & 4,876 \\
    &Gallery & Ground & 6 & 1,523  & 15,533 & 1 & 1,219 & 37,202 \\
    \multirow{2}*{$G \rightarrow A$} & Query & Ground & 6 & 1,523 & 3,046 & 1 & 1,219 & 4,876\\
    &Gallery & Aerial & 3& 1,523  & 7,717 & 1 & 1,219 & 62,791\\
    \hline
    \multirow{2}*{$G \rightarrow A+G$} & Query & Ground & 6 & 1,523 & 3,046 & - & - & - \\
    &Gallery & Aerial+Ground & 9 & 1,523  & 20,204 & - & - & - \\
    \hline
    \end{tabular}
}
\end{table*}
\section{Datasets}
To expand the datasets available for the AGPReID task, we contribute the \textbf{LAGPeR} and \textbf{G2APS-ReID} datasets as shown in Tab.~\ref{tab:dataset}. The LAGPeR dataset is independently collected, annotated, and partitioned by us, and it includes data from \textbf{21 cameras}, \textbf{7 scenes}, and \textbf{3 perspectives} (with ground perspectives divided into oblique and frontal views). For detailed information on data collection, annotation, partitioning, and experimental setup, please refer to Sec.~\ref{sec:lagper}. The G2APS-ReID dataset is reconstructed from the large-scale person search dataset G2APS~\cite{zhang2023ground}. Since the original G2APS dataset only considers retrieval tasks from ground to aerial view, which do not fully meet the requirements of the AGPReID task, we re-partition the G2APS. Detailed procedures are elaborated in Sec.~\ref{sec:g2aps}.

\subsection{LAGPeR}
\label{sec:lagper}
\noindent\textbf{Data Collection:}
Our image data is obtained using fixed Hikvision and DJI drone cameras, capturing video footage from $14$ fixed and $7$ drone cameras. The dataset encompasses $4,231$ pedestrians, totaling $63,841$ images. The drone's video footage is taken from heights ranging between $20$ to $60$ meters. The data collection process spans approximately two months and includes various scenes(teaching area, supermarket, canteen, etc.), lighting conditions(such as day or night), weather conditions (sunny or rainy), and viewing angles (straight-ahead angle, oblique angle, or high angle).

\noindent\textbf{Data Annotation:}
All our datasets are manually annotated by human annotators. The image data is sampled from the collected video footage. Fixed cameras sample one frame every $10$ frames, while drone cameras sample one frame every $24$ frames. The naming convention for person images is $\text{000X\_C0Y\_00000Z.jpg}$, where $\text{000X}$ denotes the person ID, $\text{C0Y}$ indicates the camera ID, and $\text{00000Z}$ represents the frame position in the corresponding video. The cropped images are $128 \times 256$ pixels in size.

\noindent\textbf{Data Division and Experimental Setup:}
For the LAGPeR dataset, we select 12 cameras (including 8 ground cameras and 4 drone cameras) from the first four scenes as the training set, while images from 9 cameras in the remaining three scenes are used for evaluation. To simulate real-world disturbances, we include IDs that appeare only in one camera as noise items in the dataset. After filtering, we identify $1,523$ IDs for evaluation, while the remaining $2,708$ IDs are used as the training set. During the data prepossessing stage, we aim to select the most representative images for each ID as query images. To achieve this, we employ a method based on gradient histograms and K-nearest neighbor clustering algorithms. 
Ultimately, we successfully select $3,046$ images of $1,523$ IDs from each perspective as query images, constructing a comprehensive task setup (as shown in Tab.~\ref{tab:split}).
In terms of the experimental setup, compared to conventional AGPReID datasets, we add $G \rightarrow A+G$ setting, which includes images from both ground and aerial viewpoints in the gallery to evaluate the model's performance more comprehensively.

\subsection{G2APS-ReID}
\label{sec:g2aps}
\noindent\textbf{Reconstruction and Experimental Setup:}
The original G2APS~\cite{zhang2023ground} dataset primarily focuses on ground-to-aerial retrieval and has limitations on the single evaluation perspective for the Person Search. Therefore, we reconstruct the G2APS to create the G2APS-ReID dataset. Specifically, we randomly select $60\%$ of the IDs as the training set and the remaining as the test set. We manually adjust the IDs in the test set by reallocating IDs with too few or too many images to the training set, resulting in $1,219$ IDs for evaluation and $1,509$ IDs for training. Regarding the experimental setup, since each ID in the G2APS~\cite{zhang2023ground} dataset includes only two cameras (one ground and one aerial), cross-camera retrieval cannot be performed from a ground perspective. Therefore, we do not include the $G \rightarrow A+G$ setting.

\begin{table*}[ht]
\centering
\begin{center}
   \caption{Performance comparison on LAGPeR and G2APS-ReID datasets. `$A \rightarrow G$' denotes that the Aerial view is the query, `$G \rightarrow A$' denotes that the Ground view is the query, and `$G \rightarrow A+G$' indicates that the gallery contains images from both the Aerial and Ground view. CLIP-ReID$^{*}$ indicates using OLP and SIE in Clip-ReID. MIP$^{\dagger}$ represents the re-implementation for the AGPReID. AG-ReID$^{\ddagger}$ indicates removing the attributes branch of the AG-ReID method. The best performance is in \textbf{bold}.} 
    \label{tab:gareid_result}
\end{center}
\vspace{-0.5cm}
\renewcommand\arraystretch{1.2}
\setlength{\tabcolsep}{1.5mm}{
    \begin{tabular}{lc|cc|cc|cc|cc|cc} \hline
     \multicolumn{1}{c}{\multirow{3}{*}{\textsc{Method}}} &	
     \multirow{3}{*}{\textsc{Backbone}}& \multicolumn{6}{c|}{LAGPeR} & \multicolumn{4}{c}{G2APS-ReID}\\
     
     &&\multicolumn{2}{c}{$A \rightarrow G$} & \multicolumn{2}{c}{$G \rightarrow A$} &\multicolumn{2}{c|}{$G \rightarrow A+G$} 
     & \multicolumn{2}{c}{$A \rightarrow G$} & \multicolumn{2}{c}{$G \rightarrow A$} \\ 
    \cline{3-12}
    & & Rank-1 & mAP & Rank-1	& mAP & Rank-1 &	mAP	& Rank-1 & mAP & Rank-1 & mAP \\
    \hline \hline
    ViT~\cite{luo_bag_2019} &	ViT& 38.67& 	27.25& 	32.04& 	30.69 &	18.88 &	15.31& 	69.38 &	52.17 & 67.16 &	52.22 \\
    TransReID~\cite{he2021transreid} & ViT	&38.80 &	28.80 &	33.00 &	32.10 &	22.90 &	18.80 &	67.10& 	53.10& 	68.52 &	54.19  \\

    CLIP-ReID~\cite{li2023clip}	& CLIP & 24.40& 	17.60 &	21.30 &	20.80 &	12.30& 	10.20& 	58.30 &	42.20& 	56.41	& 41.92  \\ 

    CLIP-ReID$^{*}$~\cite{li2023clip} & CLIP& 23.10& 	17.50& 	20.00& 	20.30& 	9.00& 	8.40& 	59.60 &	42.70 &	56.39 &	42.52 \\
    MIP$^{\dagger}$ ~\cite{wu2024enhancing} & ViT & 39.30 & 	29.30 & 	33.90  &	32.60  &	21.00  &	17.30 	 &73.00  &	57.40  &	70.22 &	57.06 \\ \hline
    AG-ReID$^{\ddagger}$~\cite{nguyen2023aerial} & ViT &40.48& 	28.89& 	32.96& 	31.91& 	22.03 &	17.89 	&70.75& 	52.87 &	68.70 &	53.39  \\ 
    VDT~\cite{zhang2024view} & ViT	&40.15 &	28.97& 	33.55 &	31.98 &	19.50& 	16.45& 	73.05 &	56.23 &	71.08 &	56.01  \\
    \hline
    \textbf{SeCap(Ours)} & ViT &	\textbf{41.79} 	&\textbf{30.37} &	\textbf{35.26} &	\textbf{33.42} &	\textbf{24.39} &	\textbf{19.24} &	\textbf{75.31} &	\textbf{58.57} &	\textbf{73.22} &	\textbf{58.90} \\
    \hline
    \end{tabular}
}
\end{table*}

\begin{table}[ht]
\centering
\begin{center}
   \caption{Performance comparison under two settings of AG-ReID.v1 dataset. `$A \rightarrow G$' and `$G \rightarrow A$' represent the performance in two cross-view settings. `BB' refers to the backbone. CLIP-ReID$^{*}$ indicates using OLP and SIE in Clip-ReID. The best performance is in \textbf{bold}.} 
   \label{tab:agreid_result}
\end{center}
\vspace{-0.5cm}
\renewcommand\arraystretch{1.2}
\setlength{\tabcolsep}{1mm}{
    \begin{tabular}{l cc cc c} \hline 
    \multicolumn{1}{c}{\multirow{2}*{\textsc{Method}}} &\multirow{2}{*}{\textsc{BB}}&	
         \multicolumn{2}{c}{$A \rightarrow G$}&	
         \multicolumn{2}{c}{$G \rightarrow A$}\\ 
    \cline{3-6}
    & & Rank-1 & mAP & Rank-1	& mAP \\ 
    \hline \hline
    CLIP-ReID~\cite{li2023clip} & CLIP & 72.61 & 62.09 & 74.12 & 64.19\\
    CLIP-ReID$^{*}$~\cite{li2023clip} & CLIP & 74.81 & 64.18 & 74.82& 66.11\\ 
    ViT~\cite{luo_bag_2019} & ViT &78.81&69.18 &81.61&73.03\\ \hline
    AG-ReID~\cite{nguyen2023aerial} & ViT  &81.47&72.38&82.85&73.35\\
    VDT~\cite{zhang2024view} & ViT &82.91&74.44&86.59&\textbf{78.57}\\
    \hline
    \textbf{SeCap(Ours)} & ViT  &\textbf{84.03}&\textbf{76.16}&\textbf{87.01}&78.34\\
    \hline
    \end{tabular}
}
\end{table}

\begin{table*}[ht]
\begin{center}
   \caption{The efficacy of components in SeCap is evaluated on the LAGPeR and G2APS-ReID datasets. `Baseline' represents the ReID method utilizing ViT as the backbone, `LFRM' denotes the Local Feature Refinement Module, `VDT' refers to the View Decoupling Transformer, and `PRM' means the Prompt Re-calibration Module. The best performance and best improvements are in \textbf{bold}.} 
\label{tab:xr_exp}
\end{center}
\vspace{-0.5cm}
\renewcommand\arraystretch{1.2}
\centering
\setlength{\tabcolsep}{1.1mm}{
\begin{tabular}{cl|cc|cc|cc|cc|cc} \hline 
     \multirow{3}{*}{\textsc{No.}}& \multicolumn{1}{c|}{\multirow{3}{*}{\textsc{Method}}} & \multicolumn{6}{c|}{LAGPeR} & \multicolumn{4}{c}{G2APS-ReID}\\	
     && \multicolumn{2}{c}{$A \rightarrow G$}&	
     \multicolumn{2}{c}{$G \rightarrow A$}&
     \multicolumn{2}{c|}{$G \rightarrow A+G$}&	
     \multicolumn{2}{c}{$A \rightarrow G$}&
     \multicolumn{2}{c}{$G \rightarrow A$} \\ 
    \cline{3-12}
	& & Rank-1 & mAP & Rank-1	& mAP & Rank-1 &	mAP	& Rank-1 & mAP & Rank-1 & mAP \\ 
    \hline \hline

    \multirow{2}{*}{1} & \multirow{2}{*}{Baseline (ViT)}&	\multirow{2}{*}{38.67}&	\multirow{2}{*}{27.25}&	\multirow{2}{*}{32.04}&	\multirow{2}{*}{30.69}&	\multirow{2}{*}{18.88}&	\multirow{2}{*}{15.31}&	\multirow{2}{*}{69.38}&	\multirow{2}{*}{52.17}&	\multirow{2}{*}{67.16}&	\multirow{2}{*}{52.22} \\
    &&&&&&&&&&& \\
    \hline
    
     \multirow{2}{*}{2} & \multirow{2}{*}{\quad + VDT} & 40.15 &	28.97 &	33.55 &	31.98 &	19.50 &	16.45 &	73.05 &	56.23 & 71.08 &	56.01 \\
     && \cellcolor{gray!20} +1.48 & \cellcolor{gray!20} +1.72 & \cellcolor{gray!20} +1.51 & \cellcolor{gray!20} +1.29 & \cellcolor{gray!20} +0.62 & \cellcolor{gray!20} +1.14 & \cellcolor{gray!20} +3.67 & \cellcolor{gray!20} +4.06 & \cellcolor{gray!20} +3.92 & \cellcolor{gray!20} +3.79\\
     
     \multirow{2}{*}{3} & \multirow{2}{*}{\quad + LFRM} &	40.05 &	28.76 &	33.85 &	32.10 &	22.23 &	18.45  &	72.74&	56.06&	70.47 &	56.60 \\
     && \cellcolor{gray!20} +1.38& \cellcolor{gray!20} +1.51& \cellcolor{gray!20} +1.81& \cellcolor{gray!20} +1.41 & \cellcolor{gray!20} +3.35& \cellcolor{gray!20} +3.14& \cellcolor{gray!20} +3.36& \cellcolor{gray!20} +3.89& \cellcolor{gray!20} +3.31& \cellcolor{gray!20}+4.38 \\
     \hline
    
     \multirow{2}{*}{4} & \multirow{2}{*}{\quad + LFRM + PRM}&39.36 &	28.52 &	33.06 &	31.46 &	22.13 &	17.47  &	72.72 &	55.89 &	69.79 &	56.18 \\
     && \cellcolor{gray!20} +0.69 & \cellcolor{gray!20} +1.27& \cellcolor{gray!20} +1.02& \cellcolor{gray!20} +0.77	&\cellcolor{gray!20} +3.25 & \cellcolor{gray!20} +2.16& \cellcolor{gray!20} +3.34 & \cellcolor{gray!20} +3.72& \cellcolor{gray!20} +2.63 & \cellcolor{gray!20} +3.96\\
      \multirow{2}{*}{5} & \multirow{2}{*}{\quad + LFRM + VDT} & 41.56 & 	30.15 &	34.96 & 33.38 &	23.57 	&18.96 &	73.58 &	57.42 &	72.31 &	58.02 \\
     && \cellcolor{gray!20} +2.89 & \cellcolor{gray!20} +2.90 & \cellcolor{gray!20} +2.92 &	\cellcolor{gray!20} +2.69 &	\cellcolor{gray!20} +4.69 &	\cellcolor{gray!20} +3.65 &	\cellcolor{gray!20} +4.20 &	\cellcolor{gray!20} +5.25 & \cellcolor{gray!20} +5.15 &	\cellcolor{gray!20} +5.80\\
     \hline
     
    \multirow{2}{*}{6} &\multirow{2}{*} {\quad + LFRM + VDT + PRM (\textit{Ours})}&	\textbf{41.79}& 	\textbf{30.37}& 	\textbf{35.26}& 	\textbf{33.42}& 	\textbf{24.39}& 	\textbf{19.24} &	\textbf{75.31}&	    \textbf{58.57}&	\textbf{73.22} &	\textbf{58.90} \\
    && \cellcolor{gray!20} \textbf{+3.12}&	\cellcolor{gray!20} \textbf{+3.12}&	\cellcolor{gray!20} \textbf{+3.22}&	\cellcolor{gray!20} \textbf{+2.73} & \cellcolor{gray!20} \textbf{+5.51} & \cellcolor{gray!20} \textbf{+3.93} &	\cellcolor{gray!20} \textbf{+5.93}&	\cellcolor{gray!20} \textbf{+6.40}&	\cellcolor{gray!20} \textbf{+6.06}&	\cellcolor{gray!20} \textbf{+6.68}\\
    \hline
    
\end{tabular}
}
\end{table*}

\section{Experiments}
\subsection{Experimental Settings}
\noindent\textbf{Dataset.}
We evaluate SeCap using five AGPReID datasets, including the existing AG-ReID.v1~\cite{nguyen2023aerial}, AG-ReID.v2~\cite{nguyen2024ag}, CARGO~\cite{zhang2024view}, and our proposed LAGPeR and G2APS-ReID. The results of AG-ReID.v2 and CARGO are presented in the supplementary material. 

\noindent\textbf{(1) LAGPeR:}
The dataset includes $4,231$ IDs and $63,841$ images, collected from $21$ cameras. Among these, samples from $2,708$ IDs are used for training, while images from the remaining $1,523$ IDs constitute the test set. The test set is divided into three experimental setups: $A \rightarrow G$, $G \rightarrow A$, and $G \rightarrow A+G$, with specific divisions detailed in Tab.~\ref{tab:split}. To better evaluate the model's robustness against interference, we additionally include images with incorrectly labeled IDs as noise items in the gallery. These images only appear under a single camera, and are marked as $-1$.

\noindent\textbf{(2) G2APS-ReID}
The dataset contains $200,864$ images from $2,788$ IDs, captured by two types of cameras: ground and aerial. Of these, samples from $1,509$ IDs are used for training, while samples from the remaining $1,219$ IDs are designated for testing. The test set includes two experimental setups: $A\rightarrow G$ and $G\rightarrow A$. We also include the data that appears under a single view as noise items in the gallery. Since there is only one ground camera, we do not include the $G\rightarrow A+G$ experimental setting.

\noindent\textbf{(3) AG-ReID.v1:}
The dataset comprises $21,893$ images with $388$ IDs, captured by two cameras: ground and aerial. Of these, $199$ IDs are designated for the training set, while the remaining $189$ IDs are used for the test set. The dataset also includes $15$ attributes to aid in cross-view matching. In terms of experimental setup, the test images are evaluated under two experimental settings: $A \rightarrow G$ and $G \rightarrow A$.

\noindent\textbf{Metric.}
To comprehensively evaluate SeCap, we adopt Rank-1 cumulative matching characteristics and mean Average Precision (mAP) as evaluation metrics. These metrics quantitatively assess the model’s retrieval capability from both accuracy and recall, providing strong support for subsequent model optimization and comparative analysis.

\noindent\textbf{Implementation Details.}
Our method is implemented on the PyTorch and utilizes one NVIDIA RTX 3090 GPU for all experiments. Our model employs the Vision Transformer~\cite{liu2021viT}, pre-trained on ImageNet~\cite{deng2009imagenet} as the backbone model. During inference and training, the inputs are resized to $256\times128$. In the tokenization process, the patch and stride sizes are set to $16\times16$, and the embedding shape $d$ of tokens is set to $768$. Data augmentation is applied to transform the images during training, including random cropping, color jittering, random erasing, etc. The batch size is $64$, comprising $16$ identities with $4$ images per identity. We adopt the soft version of triplet loss~\cite{ye2021deep} to avoid manual selection of $m$ in the triplet loss formulation. The model is trained $120$ epochs using the Stochastic Gradient Descent (SGD)~\cite{montavon2012neural} optimizer. A cosine learning rate decay schedule is utilized, reducing the learning rate from $8\times10^{-3}$ to a final value of $1.6\times10^{-6}$. During inference, no data augmentation or re-ranking techniques are applied. 

\subsection{Performance of SeCap}
As shown in Tab.~\ref{tab:gareid_result} and Tab.~\ref{tab:agreid_result}, for evaluation fairness, we focus exclusively on Transformer-based architectures (particularly ViT~\cite{liu2021viT}). Specifically, we compare the single-view ReID methods ViT~\cite{luo_bag_2019}, Clip-ReID~\cite{li2023clip}, and TransReID~\cite{he2021transreid}, all using ViT as the backbone (the image encoder of CLIP-ReID uses the ViT-based method). We also compare AGPReID methods, including AG-ReID~\cite{nguyen2023aerial} and VDT~\cite{zhang2024view}, as well as the cross-modal ReID method MIP~\cite{wu2024enhancing}, re-implemented for AGPReID.

\noindent \textbf{Our SeCap method can achieve state-of-the-art (SOTA) performance on the LAGPeR and G2APS-ReID datasets}. Compared to the baseline method (ViT), our SeCap demonstrates significant performance improvements. Our approach outperforms all competitors across five configurations in two datasets, especially showing notable enhancements in cross-view matching tasks.
In comparison to cross-view methods, our approach consistently surpasses AG-ReID~\cite{nguyen2023aerial} and VDT~\cite{zhang2024view} across various cross-view tasks. Notably, in the $G \rightarrow A+G$ setting, which tests the model's comprehensive performance, our method significantly outperforms other cross-view methods.
Furthermore, on the AG-ReID.v1 dataset, even though our method does not use the attributes provided by the AG-ReID.v1 dataset,  we still achieve the best results in the `$A \rightarrow G$' setting, and comparable results for the `$G \rightarrow A$' setting.
These remarkable results show our Secap can learn better view-invariant features to boost model performance.

\noindent\textbf{Compared to single-view competitors}, our method demonstrates significant improvements in the cross-view settings of LAGPeR and G2APS-ReID, surpassing the Vision Transformer (ViT) by $3\%$ on LAGPeR and $6\%$ on G2APS-ReID, respectively. Such results show our SeCap can effectively mitigate view bias in the AGPReID task. 

\noindent\textbf{Compared to cross-view competitors}, our method also exhibits better performance. It is because we adaptively re-calibrate prompts to match the current view based on the view-invariant information, significantly enhancing cross-view person re-identification performance.

\subsection{Ablation Study}
To systematically evaluate the contributions of each module in our proposed SeCap method, we design and conduct a series of ablation experiments, as shown in Tab.~\ref{tab:xr_exp}. \\
(1) The experimental results indicate that the \textbf{LFRM can significantly enhance model performance} by refining local features (\#1 vs.\#3).
(2) The VDT leverages its viewpoint decoupling capabilities to \textbf{partially eliminate the interference of viewpoint factors on feature representations} (\#1 vs.\#2). However, after adding the LFRM, \textbf{the performance of the model is further improved} (\#2 vs.\#5).
(3) Although the results of \#4 show some improvement compared to the baseline, it is \textbf{no better than using the LFRM alone} (\#3 vs.\#4). However, \textbf{the model performs best when the PRM is added to \#5}. This is because the PRM relies on the view-invariant features decoupled by the VDT to re-calibrate the prompts. Without VDT, the prompts learn incorrect view-invariant information, leading to performance degradation. \textbf{Conversely, decoupling the features allows the PRM to fully leverage the correct view-invariant information, achieving the best performance.}

\subsection{Feature visualization}
\begin{figure}[h]
\centering
\begin{subfigure}[c]{0.23\textwidth}
    \includegraphics[width=\textwidth]{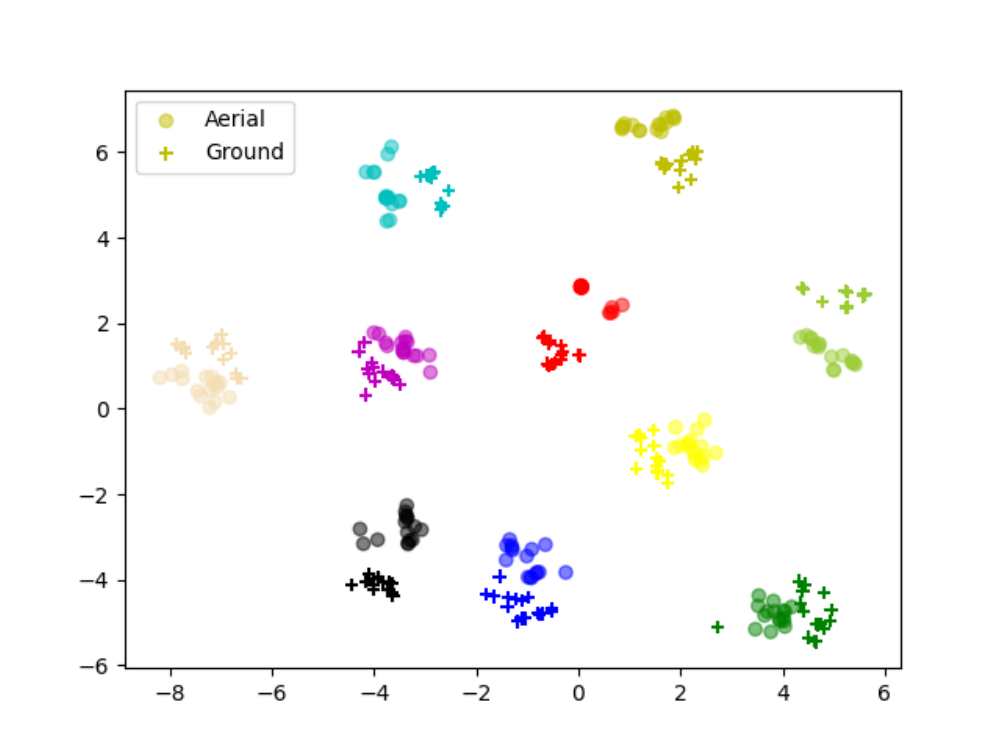}
    \caption{Baseline}
    \label{fig:agreid_len}
\end{subfigure}
\hfill
\centering
\begin{subfigure}[c]{0.23\textwidth}
    \includegraphics[width=\textwidth]{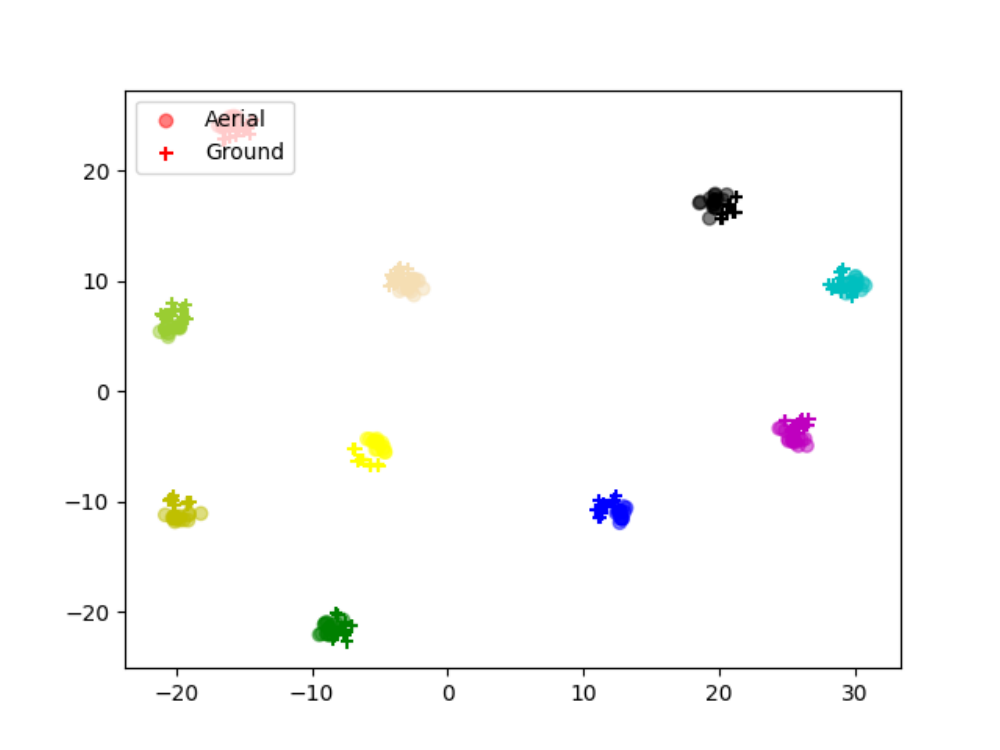}
    \caption{SeCap(Ours)}
    \label{fig:agreid_len}
\end{subfigure}
\vspace{-0.3cm}
\makeatletter\def\@captype{figure}\makeatother\caption{Visualize the features extracted by SeCap and the baseline model using t-SNE. Circles ($\bullet$) represent the Aerial View, and pluses (\textbf{+}) represent the Ground View. The same IDs are indicated by the same color.}
\label{fig:tSNE}
\vspace{-0.5cm}
\end{figure}

As shown in Fig.~\ref{fig:tSNE}, we visualize the cross-view person identity features extracted by SeCap from the LAGPeR dataset. The results demonstrate that, compared with the baseline model, our proposed SeCap exhibits stronger intra-class cohesion and inter-class discrimination. Additionally, it can effectively extract discriminative cross-view features from images with the same ID under different views.

\section{Conclusion}
This paper focuses on cross-view ReID, specifically AGPReID. Firstly, we propose the Self-Calibrating and Adaptive Prompts (SeCap) method to address the significant view differences. By re-calibrate prompts to match the current view based on the input adaptively, the SeCap significantly enhances model performance. Secondly, we contribute two real-world large-scale AGPReID datasets, LAGPeR and G2APS-ReID. Finally, the experiments on AGPReID datasets demonstrate the superiority of our method.

\noindent\textbf{Acknowledgments.}This work is supported by the National Natural Science Foundation of China (U23B2013) and the Guangdong Basic and Applied Basic Research Foundation (2025A1515011465).

{
    \small
    \bibliographystyle{ieeenat_fullname}
    \normalem
    \bibliography{main}
}
\clearpage
\setcounter{page}{1}
\maketitlesupplementary
\section{Overview}
In this supplementary material, we provide additional experimental results and more in-depth discussions of the following three aspects:
\begin{itemize}
\item We conduct a visual analysis comparing our proposed SeCap method with the baseline model, including retrieval results and attention map.
\item We perform experiments on the AGPReID datasets GARGO~\cite{zhang2024view} and AG-ReID.v2~\cite{nguyen2024ag}, demonstrating that SeCap is a feasible and effective solution for the AGPReID task across all publicly available AGPReID datasets. Additionally, we conducted cross-dataset evaluation experiments.
\item We analyze the impact of prompt length $L$ and hyperparameter $\lambda$ on model performance and conduct ablation experiments on the individual modules to verify the effectiveness of our method.
\end{itemize}

Unless otherwise specified, the numbering of figures and tables should be within the scope of the supplementary material, and consistent with the main paper.
\section{Visual Analysis}
\subsection{Retrieval Result Visualization}

\begin{figure}[ht]
\centering
\includegraphics[width=\linewidth]{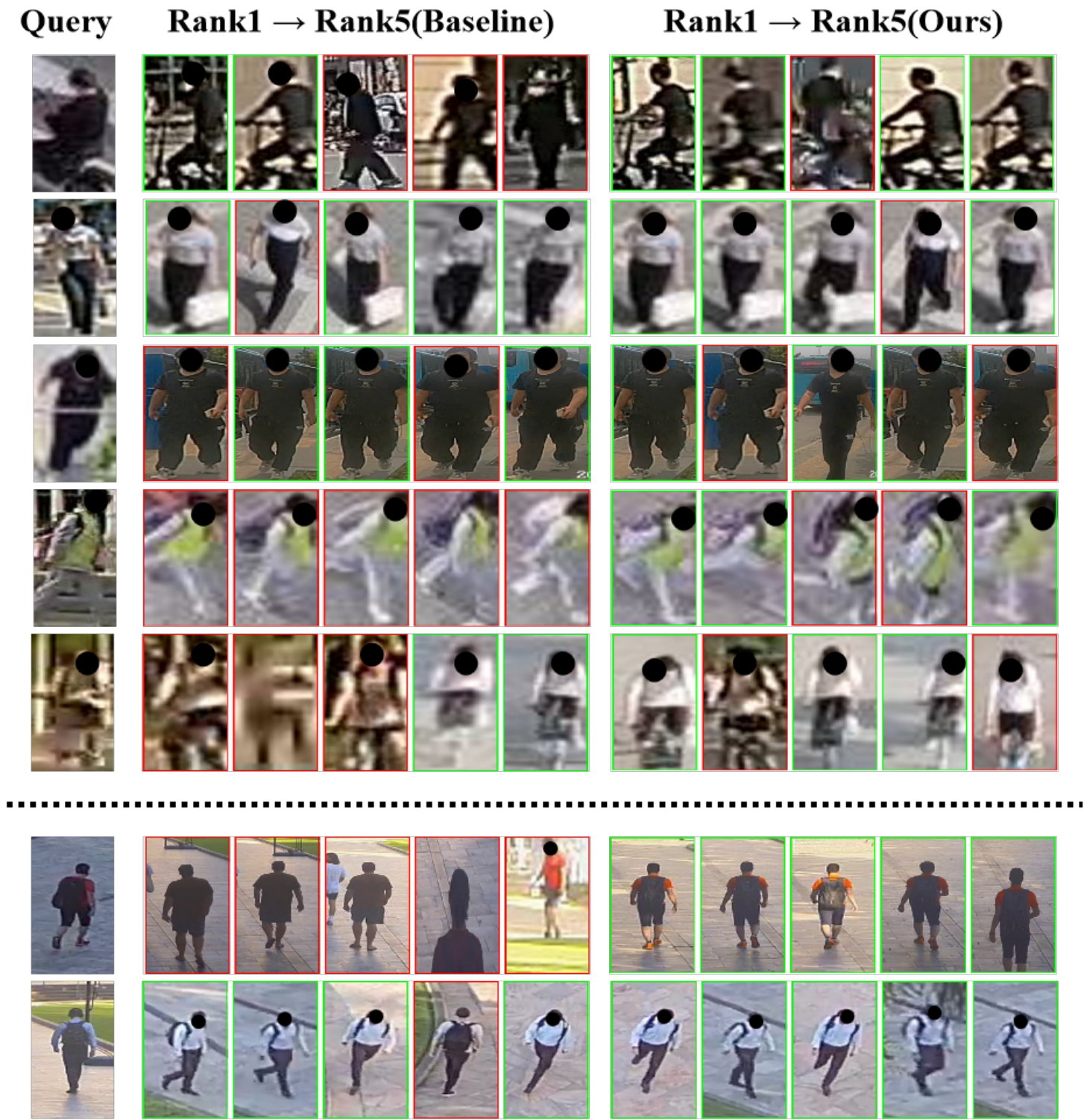}
\caption{Comparison of several retrieval visualizations on the LAGPeR dataset of setting $A\rightarrow G$. Red and green boxes represent wrong and correct matchings. The top five are listed.}
\label{fig:performance}
\end{figure}

\textbf{The visualization of the retrieval results compellingly demonstrates that SeCap is a feasible and effective method for addressing the challenges posed by AGPReID problems.} As illustrated in Fig.~\ref{fig:performance}, the retrieval outcomes on both LAGPeR and AG-ReID datasets are presented, offering a comprehensive comparison of SeCap's retrieval results with those of baseline methods across various experimental settings.

\begin{figure}[ht]
\centering
\includegraphics[width=\linewidth]{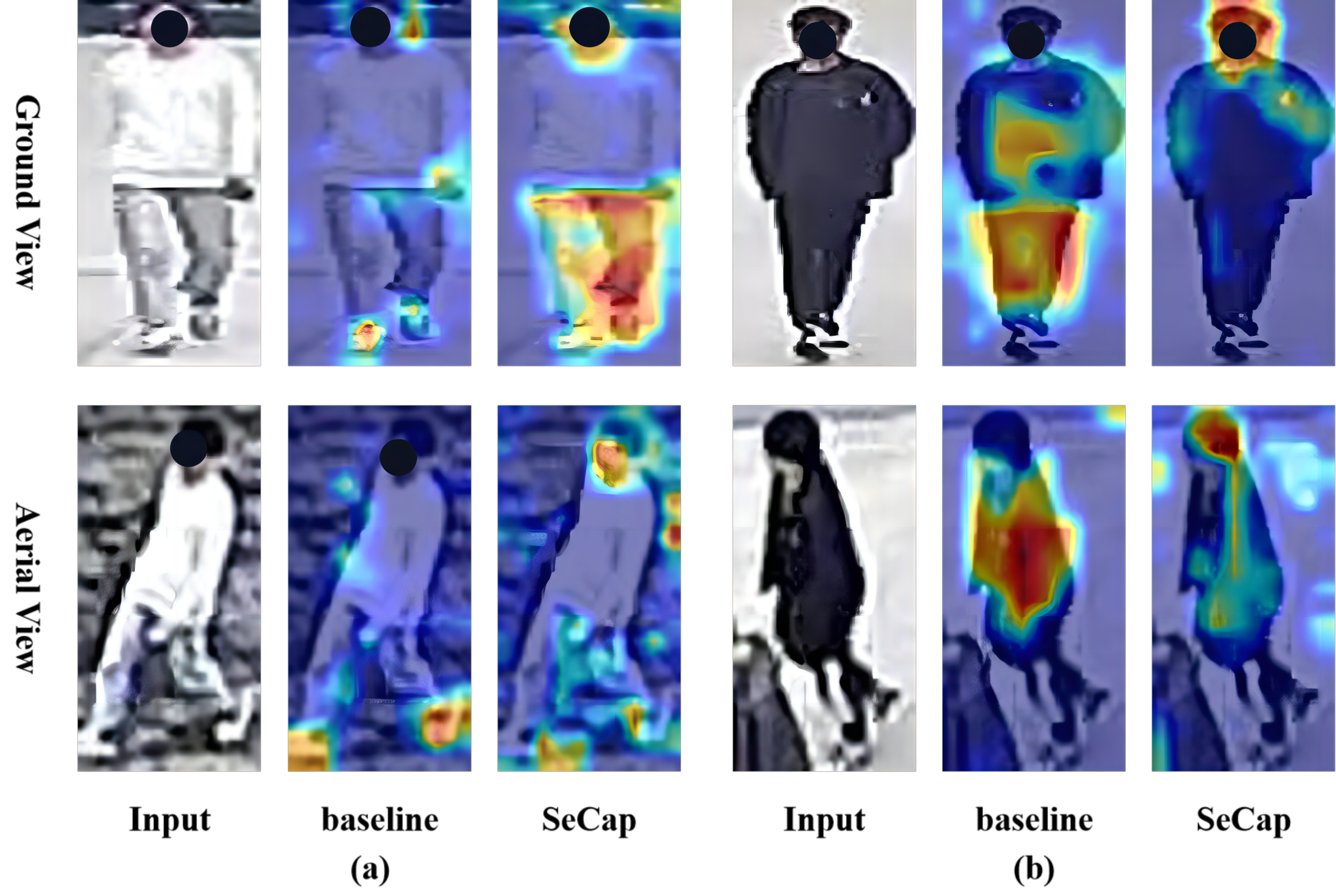}
\makeatletter\def\@captype{figure}\makeatother\caption{The visualization results of the attention maps of our SeCap method and the baseline model.}
\label{fig:attention_map}
\end{figure}

\subsection{Attention Map Visualization}
We visualized the attention maps of some case images from both the SeCap method and the baseline model. As shown in Fig.~\ref{fig:attention_map}, the baseline model tends to focus more on the torso or clothing of individuals rather than view-invariant regions like head features. \textbf{In contrast, our SeCap method effectively attends to view-invariant local features, ensuring robust performance across varying viewpoints}.

\begin{table*}[ht]
\centering
\begin{center}
   \caption{Performance comparison under CARGO dataset. `ALL' denotes the overall retrieval performance of each method. `$G\leftrightarrow G$', `$A\leftrightarrow A$', and `$A\leftrightarrow G$' represent the performance of each model in several specific retrieval patterns.Rank1 and mAP are reported (\%). The best performance is shown in \textbf{bold}.} 
\label{tab:cargo_result}
\end{center}
\vspace{-0.5cm}
\renewcommand\arraystretch{1.5}
\setlength{\tabcolsep}{3.2mm}{
    \begin{tabular}{l cc cc cc cc c} \hline 
    \multicolumn{1}{c}{\multirow{2}*{\textsc{Method}}} &\multirow{2}{*}{\textsc{BackBone}}&	
        \multicolumn{2}{c}{ALL}&
        \multicolumn{2}{c}{$G \leftrightarrow G$}&
         \multicolumn{2}{c}{$A \leftrightarrow A$}&	
         \multicolumn{2}{c}{$A \leftrightarrow G$}\\ 
    \cline{3-10}
    & & Rank-1 & mAP & Rank-1& mAP& Rank-1	& mAP &Rank-1	& mAP \\ 
    \hline \hline
     SBS~\cite{he_fastreid_2023} & R50 & 50.32& 	43.09& 	72.31 &	62.99 &	67.50& 	49.73& 	31.25 &	29.00 \\
     AGW~\cite{ye2021deep} & R50 & 60.26 &	53.44 &	81.25 &	71.66 &	67.50 &	56.48 &	43.57 	&40.90  \\
     \hline
    BoT~\cite{luo_bag_2019} & ViT & 61.54& 53.54& 82.14& 	71.34& 80.00& 	64.47& 43.13& 	40.11 \\
    VDT~\cite{zhang2024view} & ViT& 64.10& 	55.2&  	82.14& 	71.59&	\textbf{82.50}& 	66.83& 	48.12& 	42.76 \\
    \hline
    \textbf{SeCap(Ours)} & ViT&  \textbf{68.59}& 	\textbf{60.19}& \textbf{86.61}& 	\textbf{75.42}	& 	80.00& 	\textbf{68.08}& 	\textbf{69.43}& 	\textbf{58.94}  \\
    \hline
    \end{tabular}
}
\end{table*}

\begin{table*}[ht]
\begin{center}
\caption{Performance comparison on the AG-ReID.v2 dataset. C represents CCTV, W represents wearable devices, and A represents aerial views. The best results are highlighted in \textbf{bold}, while the second-best results are \underline{underlined}.} 
\label{tab:agreidv2_result}
\end{center}
\vspace{-0.5cm}
\renewcommand\arraystretch{1.5}
\setlength{\tabcolsep}{3mm}{

    \begin{tabular}{lc cc cc cc cc} \hline 
    \multicolumn{1}{c}{\multirow{2}*{\textsc{Method}}} &\multirow{2}{*}{\textsc{BackBone}}&	
         \multicolumn{2}{c}{$A \rightarrow C$}&	
         \multicolumn{2}{c}{$C \rightarrow A$}&
         \multicolumn{2}{c}{$A \rightarrow W$}&	
         \multicolumn{2}{c}{$W \rightarrow A$}\\ 
    \cline{3-10}
    & & Rank-1 & mAP & Rank-1	& mAP & Rank-1	& mAP & Rank-1	& mAP \\ 
    \hline \hline
    BoT~\cite{luo_bag_2019} & ViT & 85.40 & 77.03 & 84.65 & 75.90 & 89.77& 80.48 & 84.65 & 75.90\\ \hline
    AG-ReIDv1~\cite{nguyen2023aerial} & ViT  & 87.70 & 79.00 & 87.35 & 78.24 & \textbf{93.67} & 83.14 & \underline{87.73} & 79.08\\
    VDT~\cite{zhang2024view}	& ViT& 86.46&	79.13&	86.14&	 78.12& 90.00& 82.21& 85.26& 78.52 \\
    AG-ReIDv2~\cite{nguyen2024ag} & ViT& \textbf{88.77}&\underline{80.72}&	\underline{87.86} &	\underline{78.51} & \underline{93.62} &	\textbf{84.85} &	\textbf{88.61} &	\underline{80.11}\\
    \hline
    \textbf{SeCap(Ours)} & ViT  & \underline{88.12} & \textbf{80.84} & \textbf{88.24} & \textbf{79.99} & 91.44 & \underline{84.01} & 87.56 & \textbf{80.15}\\
    \hline
    \end{tabular}
}

\end{table*}

\section{Performance on Other AGPReID Datasets}
\subsection{Dataset.}

\noindent\textbf{(1) AG-ReID.v2:}
This dataset comprises $100,502$ images with $1,605$ unique IDs, captured by three types of cameras: CCTV, UAV, and wearable devices~\cite{nguyen2024ag}. Among these, $807$ IDs are designated for the training set, while the remaining $798$ IDs are used for the test set. Additionally, the dataset includes $15$ attributes to facilitate cross-view matching. In terms of experimental settings, the test images are evaluated under the following conditions: $A \rightarrow C$, $C \rightarrow A$, $A \rightarrow W$, and $W \rightarrow A$.

\noindent\textbf{(2) CARGO:}
The CARGO dataset is a virtual AGPReID dataset constructed using tools such as MakeHuman~\cite{briceno2019makehuman} and Unity3D. It comprises $108,563$ images with $5,000$ unique IDs, captured by $13$ cameras: $8$ ground cameras and $5$ aerial cameras. Among these, $51,451$ images from $2,500$ IDs are designated for the training set, while the remaining 51,024 images from $2,500$ IDs are used for the test set. In terms of experimental settings, the test images are evaluated under four conditions: $ALL$, $A \leftrightarrow A$, $G \leftrightarrow G$, and $A \leftrightarrow G$. The "ALL" setting focuses on comprehensive retrieval performance, while the latter targets specific retrieval scenarios.

\subsection{Performance.}

\textbf{On additional AGPReID datasets, SeCap demonstrates robust performance.} Tab.~\ref{tab:agreidv2_result} and Tab.~\ref{tab:cargo_result} present the performance of the proposed SeCap on the AG-ReID.v2~\cite{nguyen2024ag} and CARGO~\cite{zhang2024view} datasets. It can be seen that\textbf{ SeCap achieves optimal results across various settings on the synthetic AGPReID dataset CARGO and significantly outperforms other methods in the cross-view task $A\leftrightarrow G$, demonstrating the significant advantages of our proposed method in solving cross-view problems}. In the setting $A\leftrightarrow A$, due to the limited number of queries in CARGO, which consists of only 60 IDs with 134 images, the chance level is 2.5\%. Consequently, the Rank1 performance is relatively close. However, when considering the metric of mAP metric, which better reflects the model's performance, our method demonstrates superior results.

\begin{table*}[ht]
\centering
\caption{The analysis of the effectiveness of the PRM and LFRM in SeCap. LFRM stands for the Local Feature Refinement Module, PRM denotes the Prompt Re-calibration Module, and OLP represents Overlapping Patches. The meanings of Add, Cat, and Attn are detailed in the Sec.~\ref{sec:effec ana}. The best performance and the most significant improvements are highlighted in \textbf{bold}.} 
\label{tab:model analysis}
\renewcommand\arraystretch{1.5}
\setlength{\tabcolsep}{2mm}{

    \begin{tabular}{c |ccc |ccc |c |cccccc} \hline 
    \multirow{2}{*}{\textsc{No.}}& \multicolumn{3}{c|}{\textsc{PRM}} & \multicolumn{3}{c|}{\textsc{LFRM}} &\multirow{2}{*}{\textsc{OLP}} &
         \multicolumn{2}{c}{$A \rightarrow G$}&	
         \multicolumn{2}{c}{$G \rightarrow A$}&
         \multicolumn{2}{c}{$G \rightarrow A+G$}\\ 
    \cline{9-14}
    & Add & Cat & Attn. & Block & Two-Way & fusion && Rank-1 & mAP & Rank-1& mAP & Rank-1& mAP\\ 
    \hline \hline
    1   & \ding{52} & & &&\ding{52}&\ding{52}&\ding{52}& 38.48 &	27.76 &	32.14 &	30.49 &	22.75 & 18.49  \\
    2	&  & \ding{52}& &&\ding{52}&\ding{52}&\ding{52}& 40.09 &	29.10 &	33.88 &	32.71 &	22.52 &	18.44 \\
    3	&  & & \ding{52}&\ding{52}&&\ding{52}&\ding{52}& 39.92 &	28.59 &	33.32 &	31.55 &	22.88 &	18.60  \\
    4	&  & & \ding{52}&&\ding{52}&        &\ding{52}& 40.25 &	28.88 &	34.11 &	32.25 &	21.14 &	17.23   \\
    5	&  & & \ding{52}&&\ding{52}&\ding{52}&       & 40.87   &  29.11 & 33.72 & 32.48 &	19.67 &	16.40  \\
    6   &  & & \ding{52}&&\ding{52}&\ding{52}&\ding{52}& \textbf{41.8} &	\textbf{30.4} &	\textbf{35.3} &	\textbf{33.4} &\textbf{24.39} &\textbf{19.24}\\
    \hline
    \end{tabular}
}
\end{table*}

On the AG-ReID.v2 dataset, we compare AGPReID methods such as AG-ReID.v1, VDT, and AG-ReID.v2. AG-ReID.v1 only reports results using ResNet-50 as the backbone on the AG-ReID.v2 dataset, so we compare the results of ViT enhanced by the Explainable ReID Stream(EP). As shown in Tab.~\ref{tab:agreidv2_result}, we observe that \textbf{even without using the attributes provided by AG-ReID.v2, our method still achieved the best or comparable results in the $A \rightarrow C$ and $C \rightarrow A$ experimental settings. In the $A \rightarrow W$ and $W \rightarrow A$ settings, our method achieves the best or comparable mAP results, but its Rank-1 metric is not as high as AG-ReID.v2.} This discrepancy arises because our SeCap method uses view-invariant local features for matching, with head information being a significant view-invariant feature. From Fig.~\ref{fig:attention_map}, it is evident that our method implicitly trains the model to focus more on head features. Conversely, AG-ReID.v2's Elevated-View Attention Stream(EVA) explicitly uses head information for cross-view matching, which is generally more robust than implicitly extracting local features, resulting in better Rank-1 performance. However, this approach may fail when the head is occluded, leading to a significant performance drop. Therefore, our method performs better on the average mAP metric, which better indicates the model's re-identification capability~\cite{zheng2015person}. Additionally, in the $A \rightarrow W$ setting, we found that the improvement in model performance is mainly due to the attribute-based Explainable ReID Stream(EP), rather than the Elevated-View Attention Stream (EVA), which has the limitation of relying on attribute labels.

\section{Cross-dataset evaluation}
\textbf{The proposed SeCap method in this study demonstrates superiority over other methods in cross-dataset evaluation.} Specifically, as shown in Tab.~\ref{tab:cross_data}, the results of training on the LAGPeR dataset and testing on the AR-ReID dataset indicate that direct cross-dataset (or cross-domain) evaluation is a challenging task. However, the SeCap method exhibits more significant advantages compared to baseline methods and the VDT method. This advantage may stem from the dynamically generated and calibrated prompt mechanism of SeCap, which not only learns perspective-irrelevant features but also effectively guides the model to focus more on cross-domain identity discrimination features, thereby promoting the model to learn more discriminative feature representations.
\begin{table}[ht]
\centering
\caption{Cross-dataset performance evaluations (\%) for transferring from LAGPeR to AG-ReID dataset.} 
\label{tab:cross_data}
\renewcommand\arraystretch{1.5}
\setlength{\tabcolsep}{1.5mm}{

    \begin{tabular}{l cc cc c} \hline 
    \multicolumn{1}{c}{\multirow{2}*{\textsc{Method}}} &\multirow{2}{*}{\textsc{BB}}&	
         \multicolumn{2}{c}{$A \rightarrow G$}&	
         \multicolumn{2}{c}{$G \rightarrow A$}\\ 
    \cline{3-6}
    & & Rank-1 & mAP & Rank-1	& mAP \\ 
    \hline \hline
    BoT~\cite{luo_bag_2019} & ViT &33.15&	22.7&	28.90&	20.32\\
    VDT~\cite{zhang2024view}	& ViT& 34.74&	23.42&	29.83&	21.53\\

    \textbf{SeCap(Ours)} & ViT  &\textbf{37.93}&	\textbf{24.96}&	\textbf{30.87}&	\textbf{22.99}\\
    \hline
    \end{tabular}
}
\end{table}

\begin{figure*}[ht]
    \centering  
    \begin{subfigure}[c]{0.32\textwidth}
    \includegraphics[width=\textwidth]{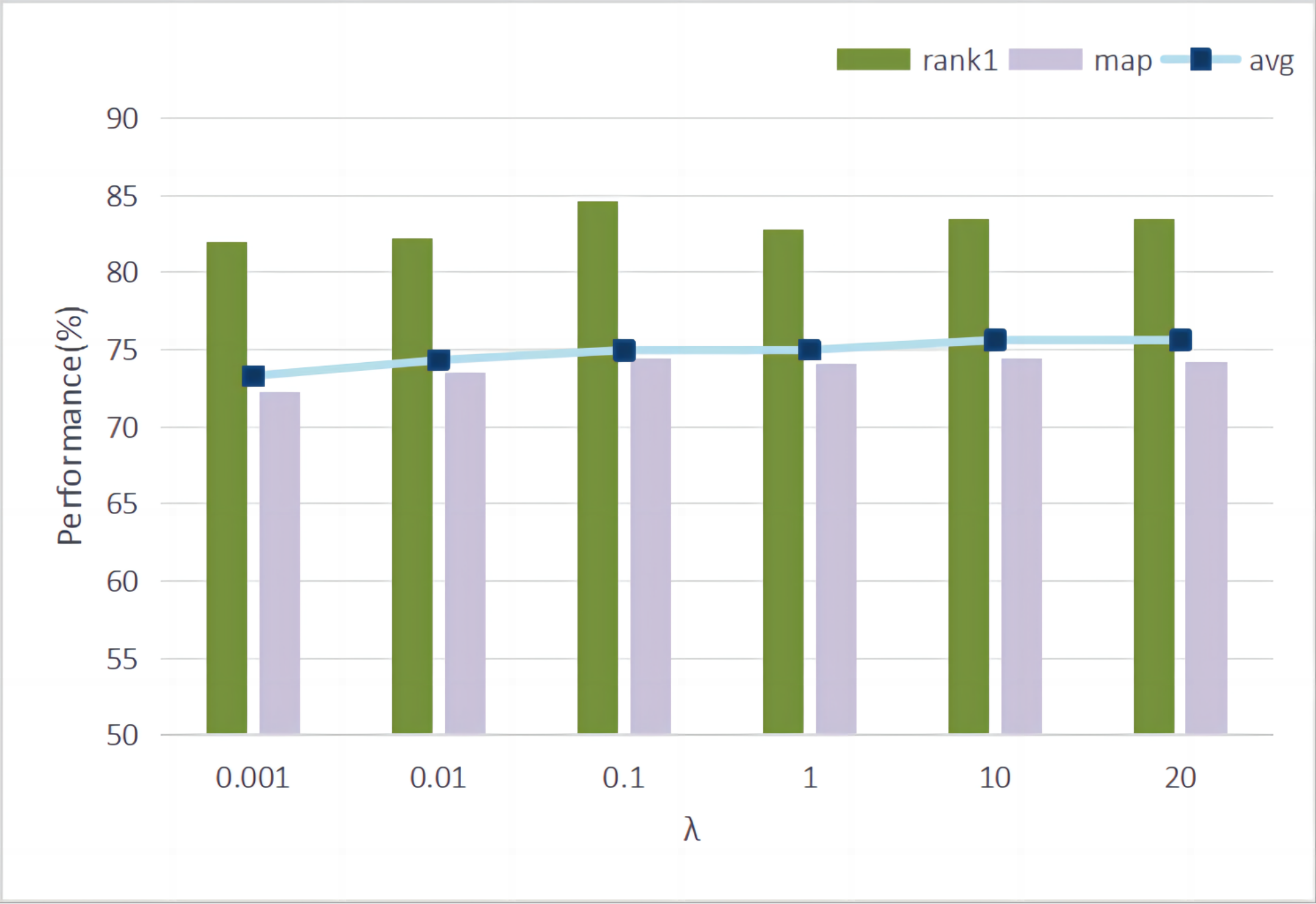}
    \caption{AG-ReID}
    \label{fig:agreid_view}
    \end{subfigure}
     \hfill
     \begin{subfigure}[c]{0.32\textwidth}
    \includegraphics[width=\textwidth]{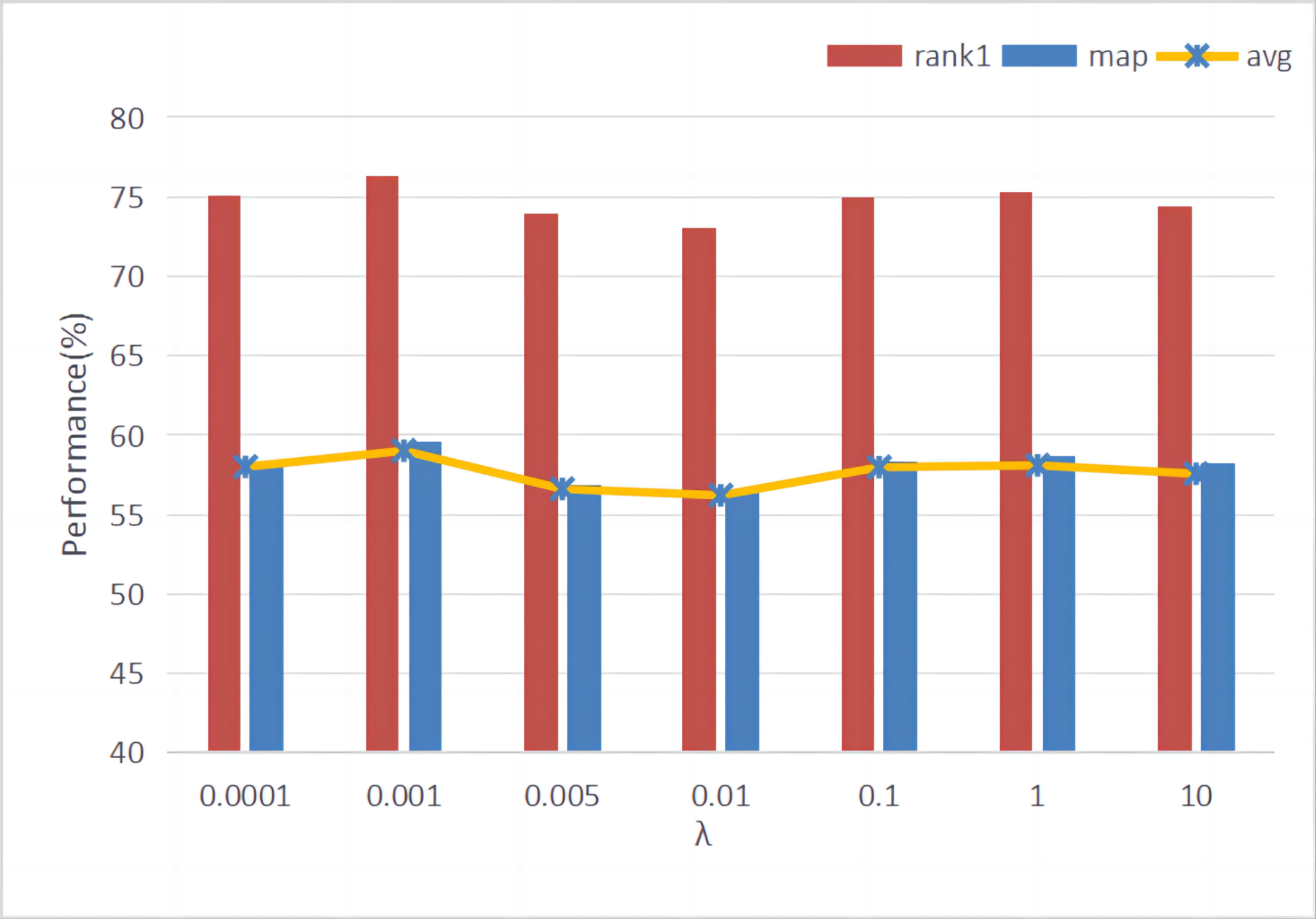}
    \caption{G2APS-ReID}
    \label{fig:g2aps_view}
    \end{subfigure}
      \hfill
    \begin{subfigure}[c]{0.32\textwidth}
    \includegraphics[width=\textwidth]{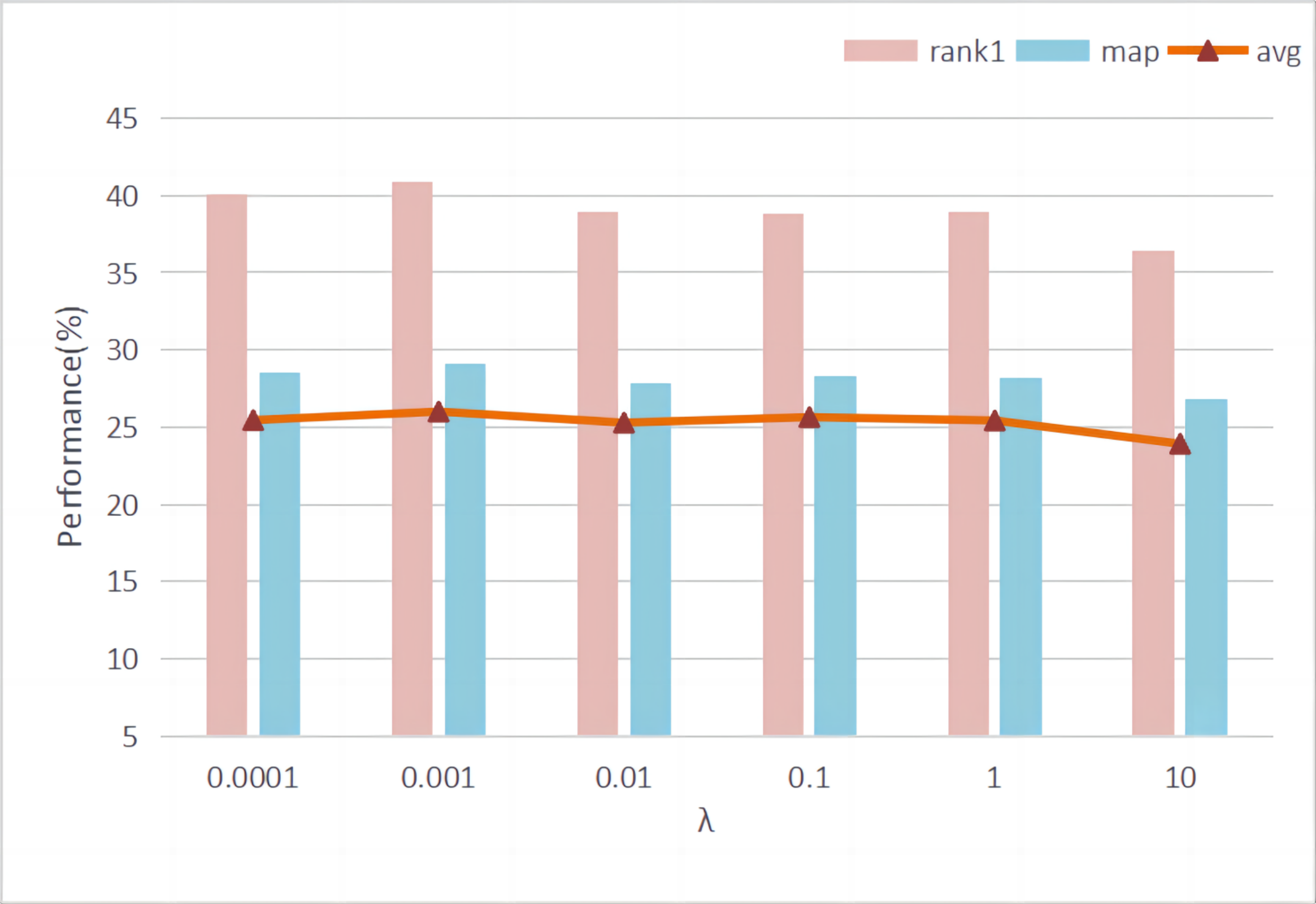}
    \caption{LAGPeR}
    \label{fig:lagper_view}
    \end{subfigure}
\caption{Fig.~\ref{fig:agreid_view} $\sim$ Fig.~\ref{fig:lagper_view} show the impact of hyperparameter $\lambda$ on model performance under three datasets. For simplicity, only setting $A\rightarrow G$ is shown on the AGPReID datasets. Rank1 and mAP are reported (\%). The avg represents the average performance of mAP.}
\label{fig:param_v}
\end{figure*}
\begin{figure*}[ht]
    \begin{subfigure}[c]{0.32\textwidth}
    \includegraphics[width=\textwidth]{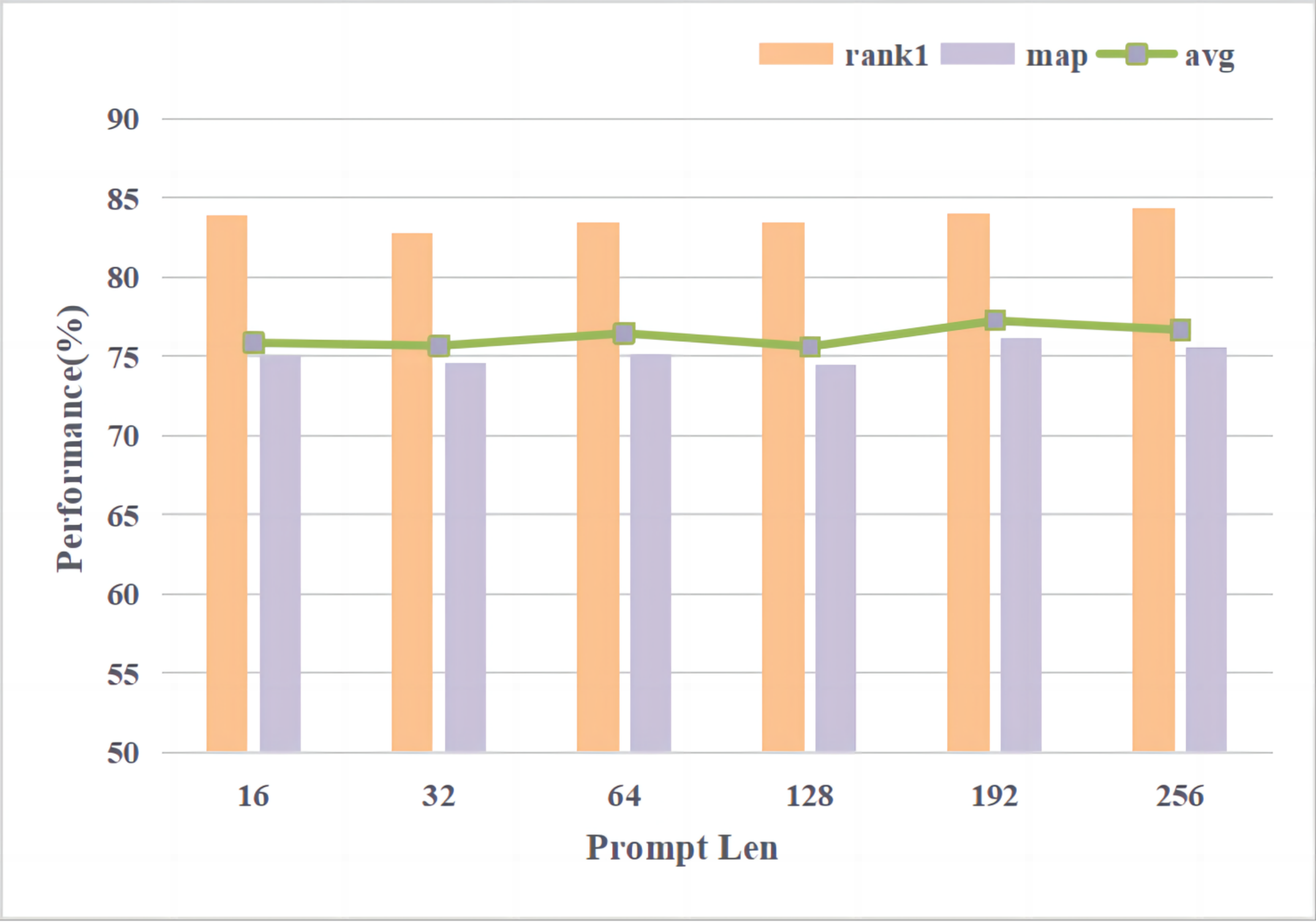}
    \caption{AG-ReID}
    \label{fig:agreid_len}
    \end{subfigure}
    \hfill
    \begin{subfigure}[c]{0.32\textwidth}
    \includegraphics[width=\textwidth]{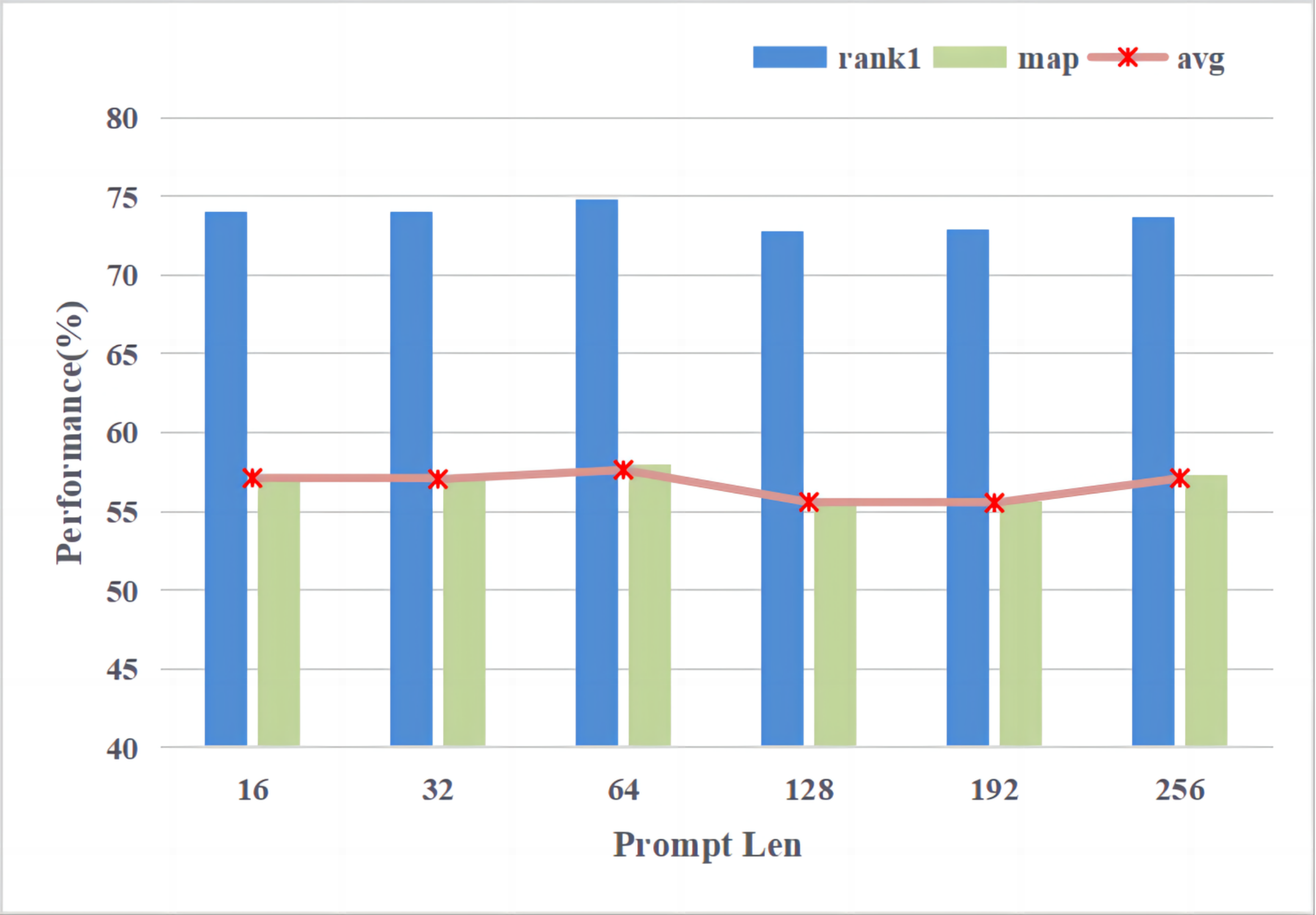}
    \caption{G2APS-ReID}
    \label{fig:g2aps_len}
    \end{subfigure}
    \hfill
    \begin{subfigure}[c]{0.32\textwidth}
    \includegraphics[width=\textwidth]{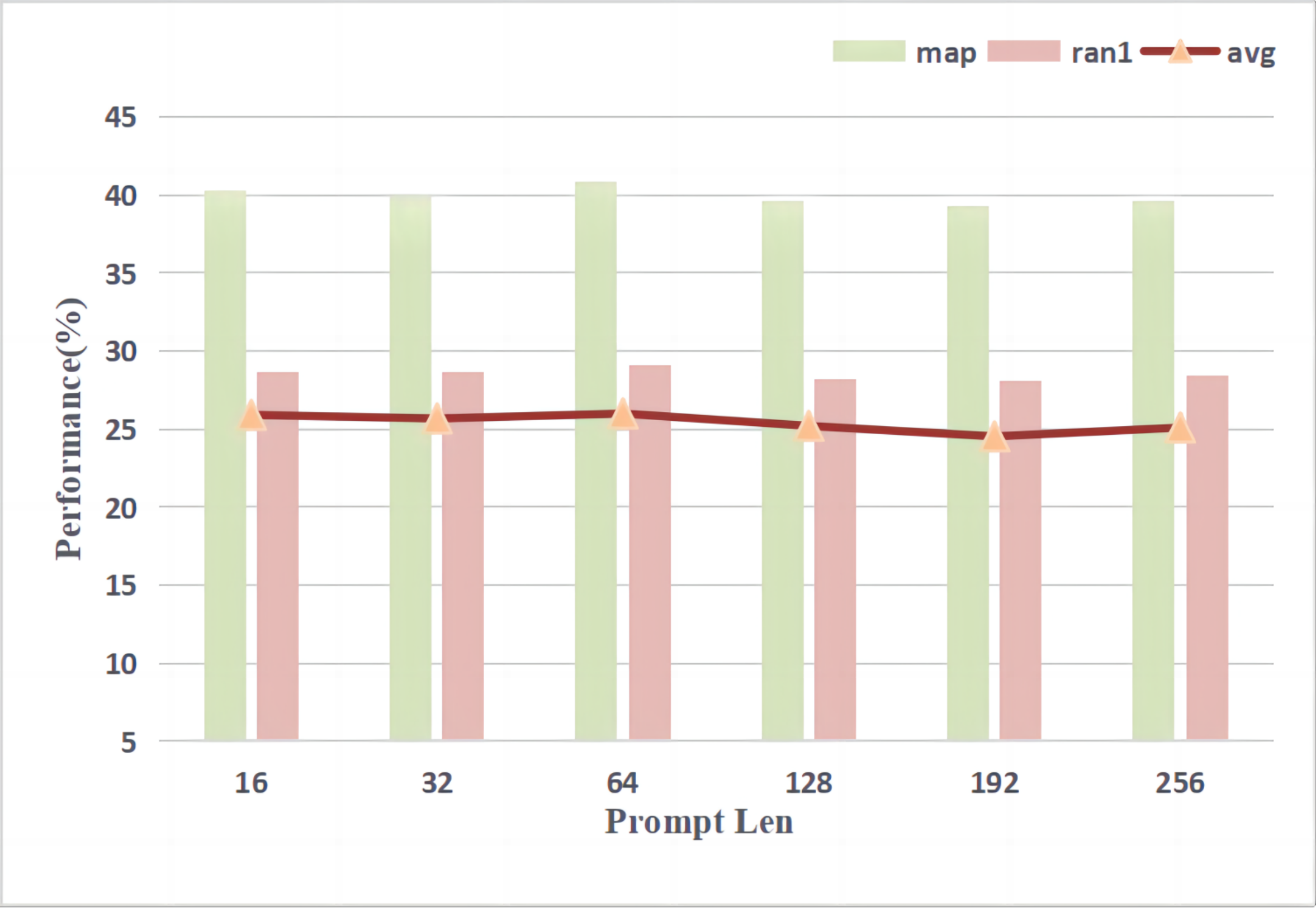}
    \caption{LAGPeR}
    \label{fig:lagper_len}
    \end{subfigure}
\caption{Fig.~\ref{fig:agreid_len} $\sim$ Fig.~\ref{fig:lagper_len} show the impact of prompt length $L$ on model performance under three datasets. For simplicity, only setting $A\rightarrow G$ is shown on the AGPReID datasets. Rank1 and mAP are reported (\%). The avg represents the average performance of mAP.}
\label{fig:param_N}
\end{figure*}

\section{Effectiveness Analysis of the Modules}
\label{sec:effec ana}
\begin{figure}[h]
\centering
\includegraphics[width=0.45\textwidth]{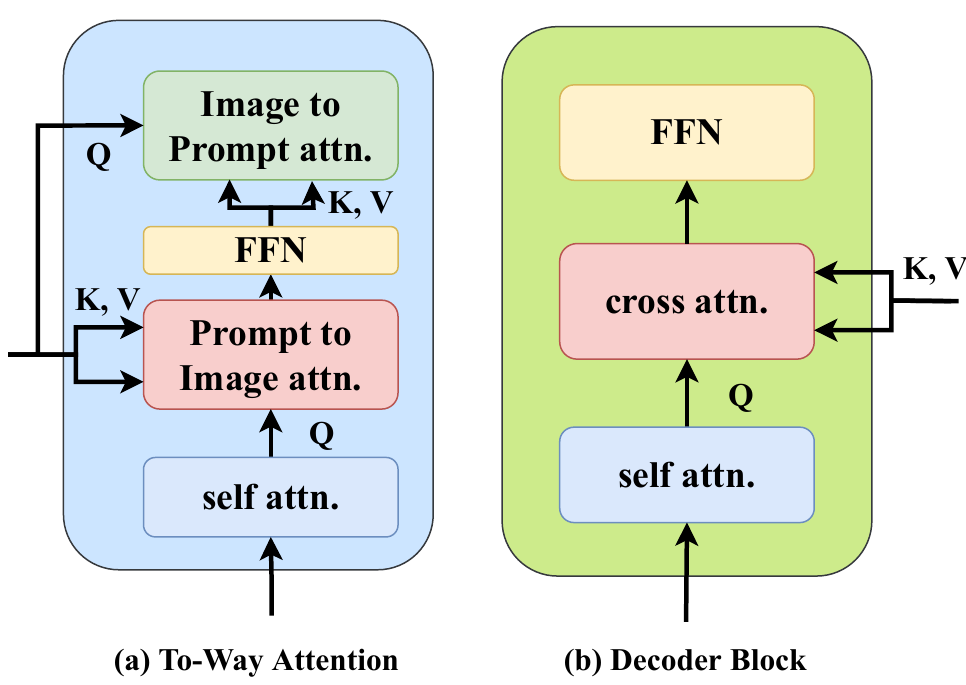}
\caption{The structure of Two-Way attention and Transformer Decoding Block.}
\label{fig:decoder}
\end{figure}

As shown in Tab.~\ref{tab:model analysis}, we analyze the roles of the Prompt Re-calibration Module (\textbf{PRM}), Local Feature Refinement Module (\textbf{LFRM}), and Overlapping Patches(\textbf{OLP}). \\
\textbf{For the Prompt Re-calibration Module(\textbf{PRM}), we explore different methods of incorporating view-invariant features by comparing the $\text{Add}$, $\text{Cat}$, and $\text{Attn}$ methods (\#1 vs \#2 vs \#6).} $\text{Add}$ represents the method of integrating view-invariant features into the prompts through addition; $\text{Cat}$ involves concatenating view-invariant features to the prompts and integrating them via self-attention; $\text{Attn}$ involves learning view-invariant information through the attention mechanism, which is the method used in \textbf{PRM}. Among these methods, $\text{Attn}$ achieves the best results. \\
\textbf{For the \textbf{LFRM} module, we compare the effects of using two-way attention and Transformer decoding blocks (\#3 vs \#6).} The two decoding structures are shown in Fig.~\ref{fig:decoder}, where the two-way attention($\text{Two-Way}$) demonstrate significant performance improvements. Additionally, we validate the effectiveness of the feature fusion module (\#4 vs \#6), confirming its utility. Lastly, we assess the impact of overlapping patches (\#5 vs \#6), which also contribute to performance enhancement.

\section{Parameter Analysis}
As illustrated in Fig.~\ref{fig:param_v}, we analyze the impact of the hyperparameter $\lambda$ on the model's performance. When $\lambda$ is set to 0.001, the SeCap model performs best on the G2APS-ReID and LAGPeR datasets. For the AG-ReID dataset, the optimal $\lambda$ is 10. \textbf{This discrepancy arises because the G2APS-ReID and LAGPeR datasets have a higher number of IDs, necessitating a smaller coefficient to balance the difficulty between viewpoint classification and ID classification.}

Under the identical $\lambda$ setting, we carry out a detailed analysis of the impact of prompt length $L$ on the model's performance. As presented in Fig.~\ref{fig:param_N}, \textbf{the model's performance is not highly sensitive to the prompt length $L$.} The model attains the best performance when the prompt length is set to $64$.

\section{Broader impact}
The proposed method can be applied to existing aerial-ground person re-identification tasks, aiming to improve the performance of AGPReID tasks. All experiments are based on publicly available datasets, reconstructed datasets from public datasets, and datasets from public datasets, with the core goal of optimizing the application effect of the recognition model in real-world scenarios, rather than deliberately designing privacy leakage mechanisms. However, it is necessary to be vigilant against potential negative effects, such as the privacy leakage risks that may arise from using surveillance and drone-captured person re-identification data. Therefore, when collecting such data, we ensure that relevant individuals are fully informed and strictly manage and use the data to protect individuals' privacy rights and interests.
\vspace{-0.5cm}

\section{Scene examples of the LAGPeR dataset.}
\begin{figure}[ht]
  \centering
  \includegraphics[width=1.0\linewidth]{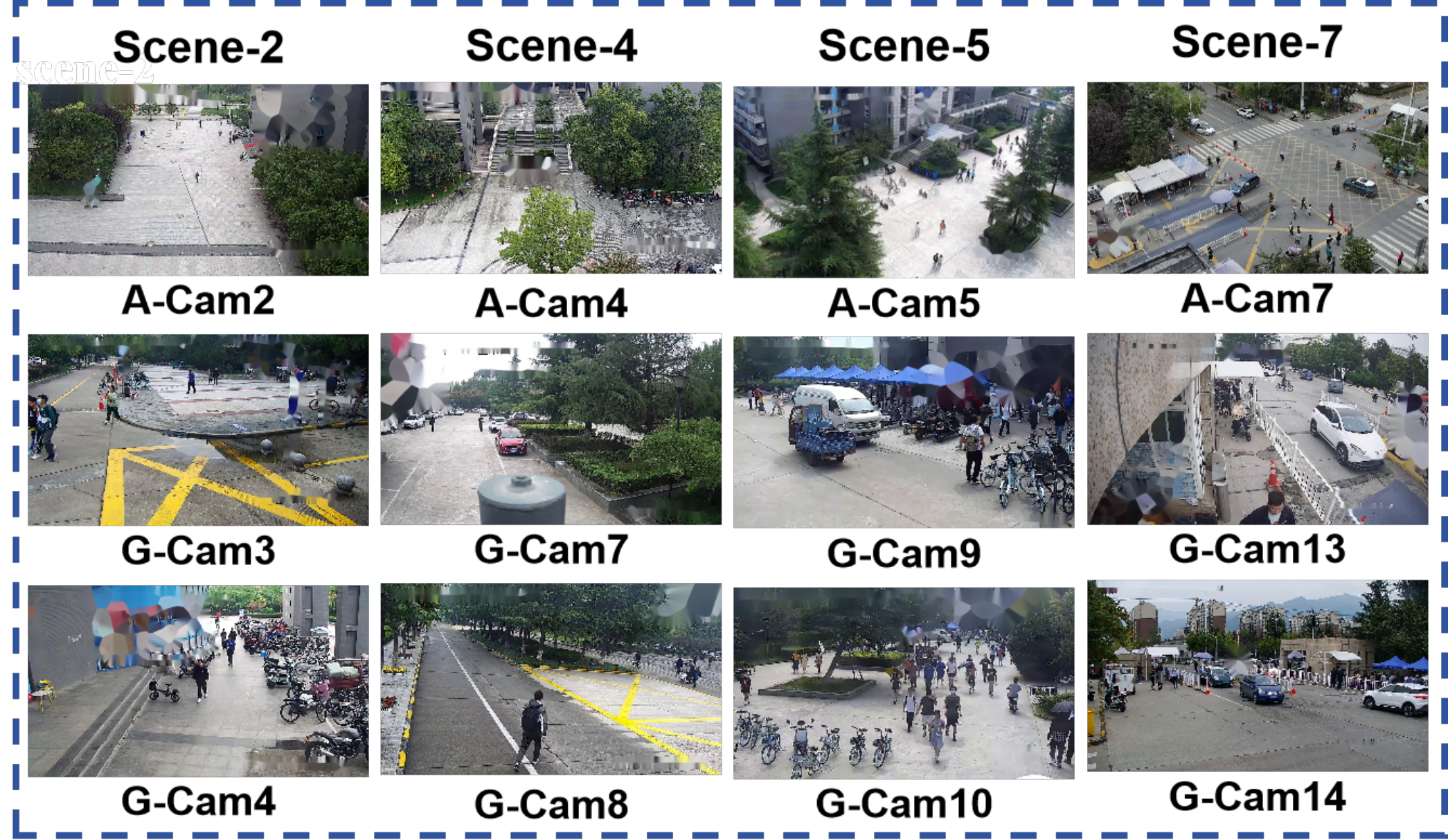}
  \vspace{-0.3cm}
   \caption{Example images from the LAGPeR, where each column represents images from the same scene.}
   \label{fig:data_scene}
\end{figure}

As Fig.\ref{fig:data_scene}, we show some scene examples of the LAGPeR dataset, where A-Cam represents the aerial view camera and G-Cam represents the ground view camera. Each column represents the same scene.
\section{Performance of $G\leftrightarrow G$ }
Our method achieves better results than top-performing compared methods in the $G$$\leftrightarrow$$G$ setting. We reconstructed the LAGPeR dataset and created the $G$$\leftrightarrow$$G$ subset to evaluate our model's performance, and the results are presented in Tab.\ref{tab:gg_result}. 

\begin{table}[ht]
\centering
\begin{center}
   \caption{Performance comparison on the $G$$\leftrightarrow$$G$ setting.} 
   \label{tab:gg_result}
\end{center}
\vspace{-0.5cm}
\renewcommand\arraystretch{1.5}
\setlength{\tabcolsep}{2.5mm}{
 \begin{tabular}{c|c cc c} \hline 
 \textsc{Method} & ViT&	VDT	&TransReID&	SeCap \\
 \hline
 Rank-1&	78.36&	77.18&	79.23&	\textbf{79.57}\\
 \hline
 mAP&	77.38&	76.35&	78.09&	\textbf{78.26} \\
 \hline
 \end{tabular}
}
\end{table}

\end{document}